\pdfoutput=1

\documentclass[11pt]{article}

\usepackage{ACL2023}

\usepackage{times}
\usepackage{latexsym}

\usepackage[T1]{fontenc}

\usepackage[utf8]{inputenc}
\usepackage{multirow}
\usepackage{multicol}
\usepackage{booktabs}
\usepackage{graphicx}
\usepackage{bbding}
\usepackage{stfloats}
\usepackage{caption}
\usepackage{subcaption}
\usepackage{soul}
\usepackage{microtype}

\newcommand{\authorspace}{\hspace{0.3cm}}

\usepackage{inconsolata}

\usepackage{booktabs}
\usepackage{graphicx}
\usepackage{supertabular}
\usepackage{xspace}
\usepackage[T1]{fontenc}
\usepackage{cleveref}

\newcommand{\model}{\textsc{ProgramFC}\xspace}

\usepackage{listings}
\usepackage{xcolor}
\usepackage{color}

\definecolor{darkred}{rgb}{0.6,0.0,0.0}
\definecolor{darkgreen}{rgb}{0,0.50,0}
\definecolor{lightblue}{rgb}{0.0,0.42,0.91}
\definecolor{orange}{rgb}{0.99,0.48,0.13}
\definecolor{grass}{rgb}{0.18,0.80,0.18}
\definecolor{pink}{rgb}{0.97,0.15,0.45}

\definecolor{codegreen}{rgb}{0,0.6,0}
\definecolor{codegray}{rgb}{0.5,0.5,0.5}
\definecolor{codepurple}{rgb}{0.58,0,0.82}
\definecolor{backcolour}{rgb}{0.95,0.95,0.92}

\lstdefinestyle{mystyle}{
  frame=single,
  basicstyle=\ttfamily\footnotesize,
  backgroundcolor=\color{backcolour}, commentstyle=\color{codegreen},
  commentstyle=\color{darkgreen}\slshape,
  keywordstyle=\color{blue},
  stringstyle=\color{darkred},
  numberstyle=\tiny\color{codegray},
  emphstyle=\color{pink}\underbar,
  morekeywords={Verify, Question},
  escapeinside={(*@}{@*)},
  breakatwhitespace=false,         
  breaklines=true,                 
  captionpos=b,                    
  keepspaces=true,                    
  numbersep=5pt,                  
  showspaces=false,                
  showstringspaces=false,
  showtabs=false,                  
  tabsize=2
}

\title{Fact-Checking Complex Claims with Program-Guided Reasoning}

\author{
  \bf Liangming Pan$^{1,2}$\authorspace
  \bf Xiaobao Wu$^3$ \authorspace
  \bf Xinyuan Lu$^{4}$ \authorspace
  \bf Anh Tuan Luu$^3$ \authorspace
      \\
  \bf William Yang Wang$^1$ \authorspace
  \bf Min-Yen Kan$^4$  \authorspace
   \bf Preslav Nakov$^2$ \vspace{2mm}
  \\
  $^1$ University of California, Santa Barbara \authorspace
  $^2$ Mohamed bin Zayed University of Artificial Intelligence \authorspace \\
  $^3$ Nanyang Technological University \authorspace
  $^4$ National University of Singapore 
  \\
  {\tt liangmingpan@ucsb.edu} \authorspace 
  {\tt xiaobao002@e.ntu.edu.sg} \authorspace
  {\tt luxinyuan@u.nus.edu} \\
  {\tt anhtuan.luu@ntu.edu.sg} \authorspace
  {\tt william@cs.ucsb.edu} \authorspace \\
  {\tt kanmy@comp.nus.edu.sg} \authorspace
  {\tt preslav.nakov@mbzuai.ac.ae}     
}

\begin{document}
\maketitle
\begin{abstract}

Fact-checking real-world claims often requires collecting multiple pieces of evidence and applying complex multi-step reasoning. In this paper, we present \textit{Program-Guided Fact-Checking} (\model), a novel fact-checking model that decomposes complex claims into simpler sub-tasks that can be solved using a shared library of specialized functions. We first leverage the in-context learning ability of large language models to generate \textit{reasoning programs} to guide the verification process. Afterward, we \textit{execute} the program by delegating each sub-task to the corresponding sub-task handler. This process makes our model both explanatory and data-efficient, providing clear explanations of its reasoning process and requiring minimal training data. We evaluate \model on two challenging fact-checking datasets and show that it outperforms seven fact-checking baselines across different settings of evidence availability, with explicit output programs that benefit human debugging.\footnote{The program code and the data are publicly available at \url{https://github.com/mbzuai-nlp/ProgramFC}}

\end{abstract}

\section{Introduction}

The proliferation of disinformation, \textit{e.g.}, in social media, has made \textit{automated fact-checking} a crucial application of natural language processing (NLP). Given a \textit{claim}, the goal is to find \textit{evidence} and then to make a \textit{verdict} about the claim's veracity based on that evidence~\cite{DBLP:conf/coling/ThorneV18,glockner-etal-2022-missing,DBLP:journals/tacl/GuoSV22}. 

\begin{figure*}[!t]
    \centering
    \includegraphics[width=16cm]{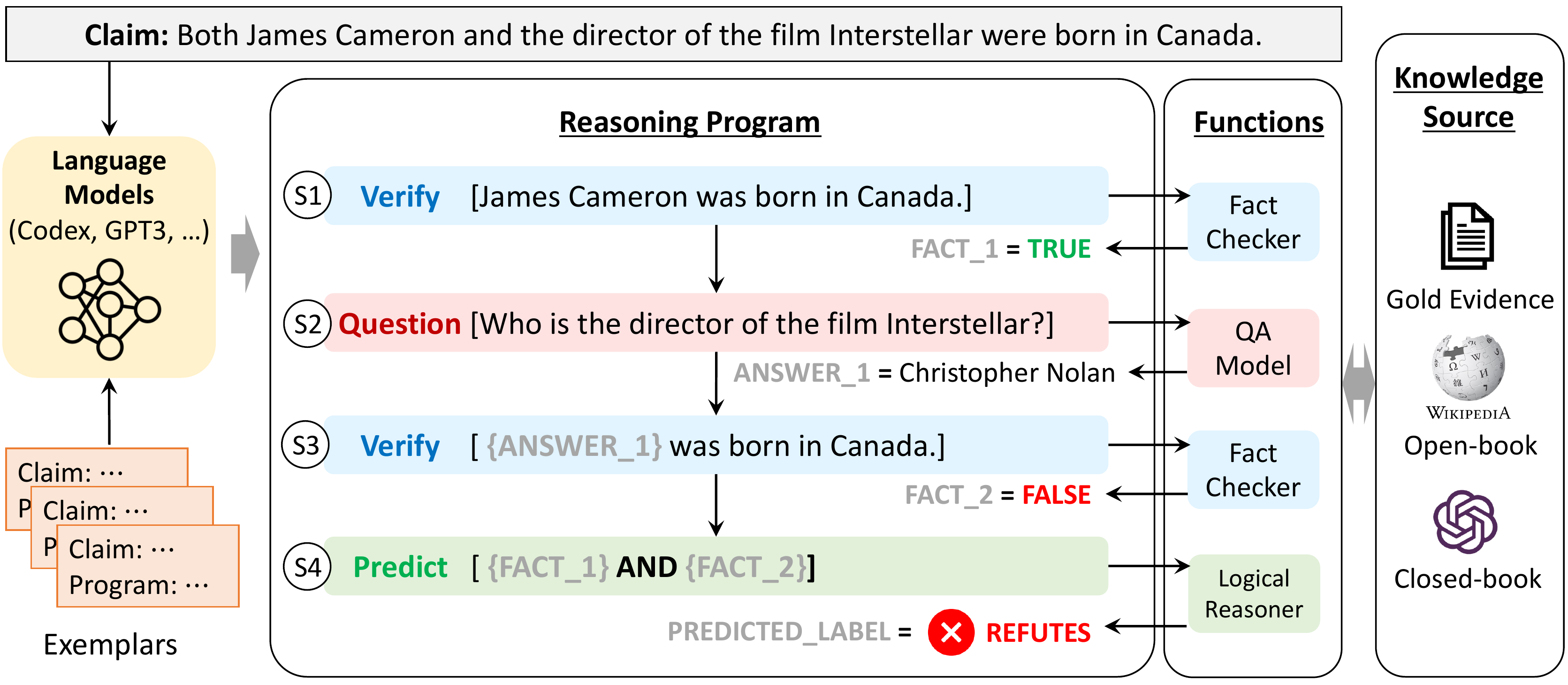}
    \caption{Overview of our \model model, which consists of two modules: (\emph{i})~\textit{Program Generation} generates a reasoning program for the input claim using Codex with in-context learning, and then (\emph{ii})~\textit{Program Execution} sequentially interprets the program by delegating each step to the corresponding sub-task function.}
    \label{fig:general_framework}
\end{figure*}

Evaluating the veracity of real-world claims often involves collecting multiple pieces of evidence and applying complex reasoning~\cite{Jiang2020HoVerAD,FANG,aly-vlachos-2022-natural,Chen2022GeneratingLA}. For instance, consider the claim ``\textit{Both James Cameron and the director of the film Interstellar were born in Canada}''. It may be challenging to find direct evidence on the web that refutes or supports this claim. Instead, a human fact-checker needs to decompose the claim, gather multiple pieces of evidence, and perform step-by-step reasoning \cite{ijcai2021p0619}, as illustrated in Figure~\ref{fig:general_framework}. This makes verifying complex claims much more challenging than the typical setting explored in previous work, where information from a single article is sufficient to support/refute the claim~\cite{DBLP:conf/naacl/ThorneVCM18,DBLP:conf/acl/SaakyanCM20,DBLP:conf/naacl/SchusterFB21,DBLP:conf/acl/PanCXKW20,wadden-etal-2022-scifact,DBLP:journals/tacl/Krishna0022}. 


Besides multi-step reasoning, we still need to consider two key aspects for developing a reliable fact-checking system: (\emph{i})~\textit{Explanability}: The model should not only predict the veracity of the claim, but it should also provide a clear explanation of its reasoning process to help users understand and trust the results. (\emph{ii})~\textit{Data efficiency}: Human annotation is often time-consuming, costly, and potentially biased, making it difficult to collect sufficient high-quality labeled data for model training, particularly for complex claims. Therefore, it is desirable to build a model that can perform well with minimal or no training data. Despite a few models~\cite{DBLP:conf/acl/ZhouHYLWLS19,DBLP:conf/acl/ZhongXTXDZWY20,aly-vlachos-2022-natural} being proposed to facilitate multi-step reasoning in fact-checking, they either lack explainability in their reasoning process or require a large number of task-specific training examples. 

In this paper, we present \textit{Program-Guided Fact-Checking} (\model), a novel fact-checking framework that is both explanatory and data-efficient. Figure~\ref{fig:general_framework} illustrates our approach. To verify complex claims, \model decomposes them into simpler sub-tasks that can be solved using a shared library of specialized sub-task functions. To be specific, \model begins by generating a \textit{reasoning program} for the input claim, which is a sequence of sub-tasks (\textit{e.g.},~S1-S4 in Figure~\ref{fig:general_framework}) in the form of \textsc{Action[Argument]}, where \textsc{Action} and \textsc{Argument} define the type and the content of the sub-task, respectively. The generated reasoning program serves as a step-by-step guide for verifying the claim. We then \textit{execute} the program by sequentially delegating each sub-task to the corresponding sub-task handler, as shown in the \emph{functions} columns in Figure~\ref{fig:general_framework}. These sub-tasks may include answering questions, verifying simple claims, or conducting logical reasoning. 

\model combines explainability with data efficiency. It uses reasoning programs to provide clear explanations of its reasoning process. For data efficiency, Large Language Models (LLMs) can solve various tasks given only a few examples as prompts, \textit{e.g.},~\textit{in-context learning}~\cite{DBLP:conf/nips/BrownMRSKDNSSAA20}. We leverage this ability of LLMs to generate reasoning programs for a given claim by showing the model just a few dozen of (claim, program) pairs as demonstrations. \model is also flexible as it allows for easy swapping of sub-task function implementations to work under different settings of fact-checking, without affecting the rest of the system. We can allow the functions to retrieve information from external sources (in an open-book setting) or we can ask them to generate answers based solely on the LLM's internal parametric knowledge (in a closed-book setting). 

We evaluate \model on two challenging datasets designed for fact-checking complex claims: HOVER~\cite{Jiang2020HoVerAD} and FEVEROUS~\cite{DBLP:conf/nips/AlyGST00CM21}, and we show that it outperforms seven few-shot fact-checking baselines on both datasets (\S~\ref{sec:main_results}). The strategy of program-guided reasoning becomes increasingly effective as the required reasoning depth increases (\S~\ref{sec:main_results}). In the open-domain setting, we find that reasoning programs can enhance the retrieval of relevant evidence from knowledge sources (\S~\ref{sec:how_reasoning_program_help}). \model is robust even when we use weak models as sub-task solvers (\S~\ref{sec:how_reasoning_program_help}). We also evaluate the interpretability of the reasoning programs through human evaluation and error analysis (\S~\ref{sec:error_analysis}). 





\section{Related Work}

\paragraph{Fact-Checking.}
Automated fact-checking has gained significant attention in the NLP research community in recent years as a means of combating misinformation and disinformation. Various datasets have been proposed that enable the development and the evaluation of systems for automatic fact-checking, the most popular ones being based on human-crafted claims from Wikipedia content~\cite{DBLP:conf/naacl/ThorneVCM18,DBLP:conf/lrec/SatheALPP20,DBLP:conf/naacl/SchusterFB21} and naturally occurring claims in the political or in the scientific domain~\cite{wang-2017liar,CheckThat:ECIR2021,augenstein-etal-2019multifc,DBLP:conf/acl/SaakyanCM20,DBLP:conf/acl/GuptaS20,DBLP:conf/emnlp/WaddenLLWZCH20,wadden-etal-2022-scifact}. Notably, most of these datasets are constructed in a way that the evidence to support or to refute a claim can be found in a \textit{single} document. For example, in FEVER~\cite{DBLP:conf/naacl/ThorneVCM18}, more than 87\% of the claims only require information from a single Wikipedia article~\cite{Jiang2020HoVerAD}. 

To bridge this gap, datasets have been proposed to study fact-checking complex claims that require multi-step reasoning~\cite{Jiang2020HoVerAD,DBLP:conf/nips/AlyGST00CM21}. Graph-based models~\cite{DBLP:conf/acl/ZhouHYLWLS19,DBLP:conf/acl/LiuXSL20,DBLP:conf/acl/ZhongXTXDZWY20,FANG,FbMultiLingMisinfo:2022,BARNABO2023100244} are used to facilitate the reasoning over multiple pieces of evidence. Although such models achieve sizable    performance gains, they lack explanability and thet rely on large amounts of training data. To address the above problems, we propose an explainable, flexible, and data-efficient model that generates reasoning graphs as explanations and utilizes in-context learning to enable few-shot learning. 

\paragraph{Explanation Generation.}
Facing the complexities of real-world claims, simply giving a final veracity to a claim often fails to be persuasive~\cite{DBLP:journals/tacl/GuoSV22}.
Previous research has proposed various approaches to provide post-hoc explanations for model predictions, such as using attention weights to highlight relevant parts of the evidence~\cite{DBLP:conf/www/PopatMSW17,DBLP:conf/cikm/CuiSW0L19,DBLP:conf/www/YangPMDYLRJH19,DBLP:conf/acl/LuL20}, generating justifications with logic-based systems based on knowledge graphs~\cite{DBLP:conf/wsdm/Gad-Elrab0UW19,DBLP:conf/tto/AhmadiLPS19}, and generating a summary of the retrieved relevant evidence~\cite{DBLP:conf/acl/AtanasovaSLA20,DBLP:conf/emnlp/KotonyaT20,DBLP:journals/information/JollyAA22}. In contrast, we propose to use reasoning programs to provide explanations that consist of sub-tasks described in a program-like natural language. This offers several advantages: it allows for explanations that are not confined to the evidence, like attention weights, it is more flexible than logic-based explanations, and it is more concise than free-form summarization. 

\paragraph{Chain-of-Thought Reasoning.}
Moreover, unlike previous work that generates post-hoc explanations, we also use reasoning programs as guidance for predicting the veracity of the claim. This is motivated by the recent success of chain-of-thought prompting (CoT)~\cite{DBLP:journals/corr/abs-2201-11903,DBLP:journals/corr/abs-2205-11916,DBLP:journals/corr/abs-2203-11171}, which generates step-by-step natural language reasoning steps to guide the model in answering complex questions. We adopt this idea to fact-checking complex claims. Unlike the original CoT, which uses a single LLM for both decomposition and question answering, we use the language model only to generate reasoning programs as the blueprint for problem-solving, and we delegate each sub-task to specialized functions. 

This approach reduces the burden on the language model and allows for more flexibility in incorporating necessary components for fact-checking such as an evidence retriever. The strategy of program-guided reasoning is also in line with the recent trend of tool-augmented language models~\cite{DBLP:journals/corr/abs-2302-07842,DBLP:journals/corr/abs-2302-04761}, \textit{i.e.}, augmenting language models with access to external tools and resources. 


\section{\model}
\label{sec:method}


We first formulate the problem of fact-checking and then we introduce our proposed model for \textit{Program-Guided Fact-Checking} (\model). 

\subsection{Problem Formulation}
\label{sec:problem_formulation}
Given a \textit{claim} $C$, a \textit{fact-checking model} $\mathcal{F}$ aims to predict a \textit{label} $Y$ to evaluate the claim as \textsc{true} or \textsc{false}, based on a \textit{knowledge source} $\mathcal{K}$. The model is also required to output an \textit{explanation} $E$ to justify the predicted veracity label. We summarize three different settings of fact-checking depending on the type of knowledge source $\mathcal{K}$. 

\vspace{0.1cm}

\noindent $\bullet$ \textbf{Gold evidence}: For each claim, $\mathcal{K}$ is the set of gold evidence documents that can support or refute the claim. This setting is also called \textit{claim verification}~\cite{DBLP:conf/acl/PanCXKW20,DBLP:conf/acl/0001WLKCAW22}. 

\noindent $\bullet$ \textbf{Open-book setting}: $\mathcal{K}$ is a large textual corpus such as Wikipedia. The model first retrieves relevant \textit{evidence} from the corpus and then predicts the veracity label based on the evidence~\cite{DBLP:conf/acl/JiangPL20,DBLP:conf/naacl/WaddenLWCBH22}. 

\noindent $\bullet$ \textbf{Closed-book setting}: The model does not have access to any external knowledge source ($\mathcal{K} = \emptyset$). It needs to leverage the knowledge stored in its parameters (acquired during pre-training and fine-tuning) to verify the claim. This setting was explored in work that applies large language models for fact-checking~\cite{DBLP:LM_as_fact_checker,DBLP:conf/naacl/LeeBMF21}. 

\subsection{Program-Guided Reasoning}
Our goal is to fact-check a complex claim $C$ that requires multi-step reasoning. We focus on the \textit{few-shot} setting, where only a small set of in-domain examples are available to teach the model. To solve this, \model follows a \textit{program generation-and-execution} paradigm, as shown in Figure~\ref{fig:general_framework}. 

\paragraph{Program Generation.} 
At this stage, given the input claim $C$, a \textit{planner} $\mathcal{P}$ generates a \textit{reasoning program} $P = [S_1, \cdots, S_n]$ for it, which consists of $n$ sequentially ordered \textit{reasoning steps} $S_i$. 

Each \textit{reasoning step} $S_i \in P$ is an instruction in controlled natural language that directs $S_i$ to a function in an auxiliary set of sub-task functions $\mathcal{F}$ available to the system. To be specific, we define $S_i = (f_i, A_i, V_i)$, where $f_i$ specifies the sub-task function $f_i \in \mathcal{F}$, $A_i$ is the \textit{argument} passed to the function $f_i$, and $V_i$ is the \textit{variable} that stores the returned result from the function call $f_i(A_i)$. For a valid reasoning program, the return value of the last reasoning step must be a Boolean value indicating the veracity label of the claim $C$, \textit{i.e.},~$V_n \in \{\textsc{True}, \textsc{False} \}$. 

\paragraph{Program Execution.}
In the execution stage, the reasoning program $P$ is run by an \textit{interpreter} to derive the veracity label of the claim $C$. The interpreter sequentially parses the reasoning steps in $P$. For each step $S_i = (f_i, A_i, V_i)$, it calls the corresponding off-the-shelf \textit{sub-task function} $f_i$ and passes the argument $A_i$ to it. The argument $A_i$ is either a logical expression or a natural language sentence, \textit{e.g.},~a question or a simple claim. The result of the function call is then stored in the variable $V_i$. As it is common for a subsequent step to depend on the results from previous steps, we allow the argument $A_i$ to \textit{refer to} variables $V_1, \cdots, V_{i-1}$ in previous steps. For example, in Figure~\ref{fig:general_framework}, the argument in $S_3$ is ``\textit{\{ANSWER\_1\} was born in Canada.}'', which refers to the return variable \textit{\{ANSWER\_1\}} from $S_2$. When executing $S_3$, the variable is replaced by its actual value, and the argument becomes ``\textit{Christopher Nolan was born in Canada}''. After executing the last step, the return value is the predicted veracity of the claim $C$. 

\paragraph{Aggregating Reasoning Paths.}
Note that there might be multiple reasoning paths that can reach the final veracity label. Therefore, we generate a diverse set of $N$ candidate reasoning programs $\mathcal{P} = \{P_1, \cdots, P_N \}$ for the input claim. After executing all programs in $\mathcal{P}$, we take the majority vote over all $N$ predicted labels as the final label. This approach is similar to how humans rely on multiple methods of validation to increase their confidence in fact-checking. It also makes the model less susceptible to errors in individual reasoning programs. 

\begin{figure*}[t]
\lstset{style=mystyle,
        frame=none,
        backgroundcolor=\color{white},
        xleftmargin=0.05\textwidth,
        xrightmargin=0.05\textwidth}
\begin{lstlisting}[language=Python]
'''Generate a python-like program that describes the reasoning steps required to verify the claim step-by-step. You can call three functions in the program: 1. Question() to answer a question; 2. Verify() to verify a simple claim; 3. Predict() to predict the veracity label.'''

# The claim is that Both James Cameron and the director of the film Interstellar were born in Canada.
def program():
    fact_1 = Verify("James Cameron was born in Canada.")
    Answer_1 = Question("Who is the director of the film Interstellar?")
    fact_2 = Verify("{Answer_1} was born in Canada.")
    label = Predict(fact_1 and fact_2)

(*@\color{codegray}{\textbf{($\cdots$ more in-context examples here $\cdots$)}}@*)

# The claim is that (*@{\color{codepurple}{\textsc{<input\_claim>}}@*)
def program():
\end{lstlisting}
\vspace{-0.2cm}
\caption{The Codex prompt template used to generate reasoning programs, consisting of a task instruction, in-context examples, and a prompt for the \texttt{<input\_claim>}. The full templates are given in Appendix~\ref{appendix:programs}. }
\label{fig:program_generation_example}
\vspace{-0.2cm}
\end{figure*}

\subsection{Reasoning Program Generation}

We base our program generator on \textit{Codex}~\cite{DBLP:journals/corr/abs-2107-03374}, a code-pretrained LLM, which can parse natural language into symbolic representations such as SQL~\cite{DBLP:journals/corr/abs-2210-02875} or Python programs~\cite{DBLP:journals/corr/abs-2211-10435,DBLP:journals/corr/abs-2211-12588}. 

However, the grammar of a reasoning program is different from the grammar of a programming language. We take advantage of Codex's few-shot generalization ability and we find that it can learn effectively from only a small number of in-context examples $\mathcal{D} = \{ d_1, \cdots, d_{|D|}\}$. Each example $d_i$ consists of a claim and a program. The program has a Python-like grammar, where each reasoning step is written in the format \textsc{$V_i = f_i(A_i)$}. At inference time, we prompt Codex with an instruction of the task, $K$ in-context examples, and the input claim $C$. Codex then attempts to complete the following texts, and thereby generates a program for $C$. The prompt template is shown in Figure~\ref{fig:program_generation_example}. We use $K = 20$ to maintain a tradeoff between the diversity of reasoning types and the model's maximum input capacity. We use sampling-based decoding (temperature of 0.7) to generate different reasoning programs for multiple runs.

\begin{figure}[!t]
	\centering
	\includegraphics[width=7.5cm]{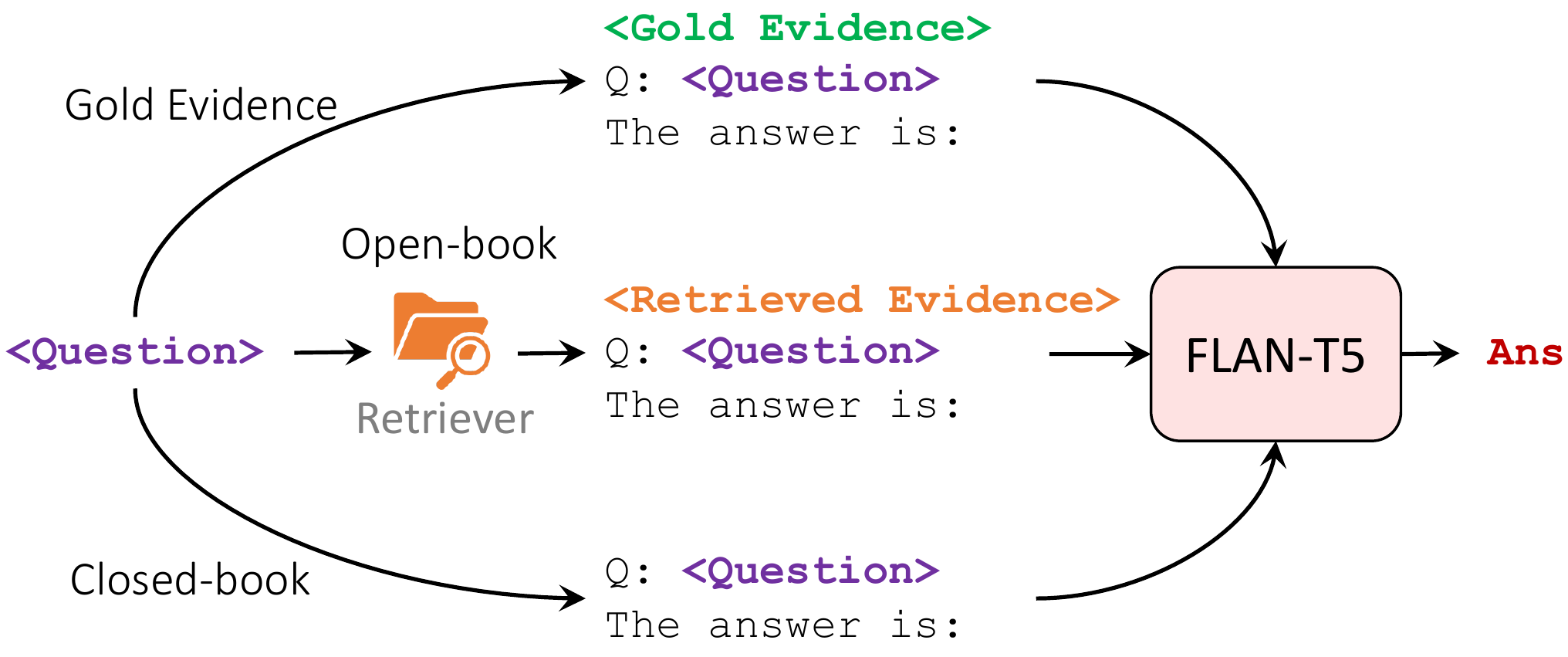}
    \caption{Implementation of the question-answering sub-task function for three different settings.}
    \label{fig:qa_module}
    \vspace{-0.2cm}
\end{figure}

\subsection{Sub-Task Functions}
\label{sec:substask_functions}

We implement three sub-task functions for the model to call during the program execution. 

\vspace{0.1cm}

\noindent $\bullet$ \textbf{\textsc{Question}:}
This sub-task function is a question-answering module that takes a question $Q$ as the input argument and returns the answer $A$ to the question. We use \texttt{FLAN-T5}~\cite{DBLP:journals/corr/abs-2210-11416}, an improved T5 model~\cite{DBLP:journals/jmlr/RaffelSRLNMZLL20} pretrained on more than 1.8K tasks with instruction tuning, which has achieved state-of-the-art zero/few-shot performance on many QA benchmarks. As shown in Figure~\ref{fig:qa_module}, we prompt the model differently depending on the settings defined in Section~\ref{sec:problem_formulation}. For the closed-book setting, the input prompt is 
\begin{quote}
\texttt{Q: \colorbox{gray}{\color{white}{QUESTION}}? The answer is:}
\end{quote}
For the other two settings, the input prompt is 
\begin{quote}
\texttt{\colorbox{gray}{\color{white}{EVIDENCE}} Q: \colorbox{gray}{\color{white}{QUESTION}}? \\ The answer is:}
\end{quote}
\noindent $\bullet$ \textbf{\textsc{Verify}:}
This is a fact verification module that takes a claim $C$ as the input argument and returns a label of either \textsc{True} or \textsc{False}. We also use \texttt{FLAN-T5} for this module, by prompting the model with the following question-answering format. 
\begin{quote}
\texttt{\colorbox{gray}{\color{white}{EVIDENCE}}
\\
Q: Is it true that \colorbox{gray}{\color{white}{CLAIM}}?
\\
True or False? The answer is:}
\end{quote}
\noindent $\bullet$ \textbf{\textsc{Predict}:}
This module takes as input a logical expression that performs \textsc{and}, \textsc{or}, \textsc{not} operations over the variables in the previous steps. Its output is returned as the predicted veracity label. 



\section{Experiments}

\paragraph{Datasets.}
Most fact-checking datasets consist primarily of simple claims that can be substantiated through a single piece of evidence. However, here we focus on complex claims that need multi-step reasoning. Given this context, we opt to evaluate our model on the only two datasets that, to the best of our knowledge, fulfill these criteria: HOVER~\cite{Jiang2020HoVerAD} and FEVEROUS~\cite{DBLP:conf/nips/AlyGST00CM21}. We use the validation sets for evaluation since the test sets are not publicly released. HOVER contains claims that require integration and reasoning over multiple Wikipedia articles. We divide its validation set into three subsets based on the number of ``hops'' required to verify the claim: 1,126 two-hop claims, 1,835 three-hop claims, and 1,039 four-hop claims. FEVEROUS focuses on fact-checking complex claims over unstructured and structured data, where each claim is annotated with evidence in the form of sentences and/or cells from tables in Wikipedia. Since we focus on textual fact-checking, we only selected claims that require exclusively sentence evidence, constituting 2,962 claims. We call this subset FEVEROUS-S. 

For evaluation in the open-book setting, we use the corresponding Wikipedia corpus constructed for these two datasets as the knowledge sources. HOVER uses the October 2017 Wikipedia dump processed by~\citet{DBLP:conf/emnlp/Yang0ZBCSM18}, consisting of the introductory sections of 5.2 million Wikipedia pages. FEVEROUS uses the December 2020 dump, including 5.4 million full Wikipedia articles. 

\paragraph{Baselines.}
We compare \model to seven baselines, categorized into three groups. (\emph{i})~\textit{Pretrained models}: \texttt{BERT-FC}~\cite{DBLP:conf/ecir/SoleimaniMW20} and \texttt{LisT5}~\cite{DBLP:conf/acl/JiangPL20} are two models that leverage BERT and T5 for fact verification, respectively. (\emph{ii})~\textit{FC/NLI fine-tuned models}: we choose three pretrained models that are fine-tuned on other fact-checking datasets or natural language inference (NLI) datasets. \texttt{RoBERTa-NLI}~\cite{DBLP:conf/acl/NieWDBWK20} uses fine-tuned RoBERTa-large on four NLI datasets; \texttt{DeBERTaV3-NLI}~\cite{DBLP:journals/corr/abs-2111-09543} fine-tunes the DeBERTaV3 model on 885,242 (claim, evidence, label) annotations from FEVER and four NLI datasets. \texttt{MULTIVERS}~\cite{DBLP:conf/naacl/WaddenLWCBH22} is a LongFormer~\cite{DBLP:journals/corr/abs-2004-05150} model fine-tuned on FEVER. (\emph{iii})~\textit{In-context learning models}: one baseline is that we directly use the \texttt{FLAN-T5} model in our \textsc{VERIFY} module for fact-checking. The other baseline uses the in-context learning of \texttt{Codex} for few-shot fact-checking. The implementation details are given in Appendix~\ref{appendix:baselines}. 

\paragraph{Few-Shot Learning.} 
We study few-shot learning where only a few in-domain examples are available. Therefore, for a fair comparison, we restrict all models to have access to only 20 examples from HOVER or FEVEROUS-S. 

We use these examples either for fine-tuning pre-trained models (\texttt{BERT-FC} and \texttt{LisT5}), for continuous fine-tuning the FC/NLI fine-tuned models, or as in-context examples for \texttt{FLAN-T5} and \texttt{Codex}. For \model, we use them as in-context examples for reasoning program generation. 

We evaluate both the \textit{gold evidence setting} and the \textit{open-book setting}. The baseline models are the same for both settings. 
However, during testing in the open-book setting, the models are given the retrieved evidence rather than the ground-truth evidence. 
We use BM25~\cite{DBLP:journals/ftir/RobertsonZ09} implemented with the Pyserini toolkit~\cite{DBLP:conf/sigir/LinMLYPN21} as the retriever for both \model and the baselines. We use as evidence the top-10 paragraphs retrieved from the knowledge corpus. 


\begin{table*}[!t]
\centering
\resizebox{\textwidth}{!}{
\renewcommand{\arraystretch}{1.05}
\begin{tabular}{ll|cc|cc|cc|cc}
\toprule
  \multicolumn{2}{l|}{\multirow{2}{*}{\textbf{Few-shot learning models}}} &
  \multicolumn{2}{c}{\textbf{HOVER (2-hop)}} &
  \multicolumn{2}{c}{\textbf{HOVER (3-hop)}} &
  \multicolumn{2}{c}{\textbf{HOVER (4-hop)}} &
  \multicolumn{2}{c}{\textbf{FEVEROUS-S}} \\ 
  \cmidrule(lr){3-4} \cmidrule(lr){5-6} \cmidrule(lr){7-8} \cmidrule(lr){9-10}
  & & \textbf{Gold} & \textbf{Open} & \textbf{Gold} & \textbf{Open} & \textbf{Gold} & \textbf{Open} & \textbf{Gold} & \textbf{Open} \\ \hline
\multirow{2}{*}{\uppercase\expandafter{\romannumeral1}} & \texttt{BERT-FC}\small{~\cite{DBLP:conf/ecir/SoleimaniMW20}} & 53.40 & 50.68 & 50.90 & 49.86 & 50.86 & 48.57 & 74.71 & 51.67 \\
& \texttt{LisT5}\small{~\cite{DBLP:conf/acl/JiangPL20}} & 56.15 & 52.56 & 53.76 & 51.89 & 51.67 & 50.46 & 77.88 & 54.15 \\ 
\midrule
\multirow{3}{*}{\uppercase\expandafter{\romannumeral2}} & \texttt{RoBERTa-NLI}\small{~\cite{DBLP:conf/acl/NieWDBWK20}} & 74.62 & 63.62 & 62.23 & 53.99 & 57.98 & 52.40 & 88.28 & 57.80 \\
& \texttt{DeBERTaV3-NLI}\small{~\cite{DBLP:journals/corr/abs-2111-09543}} & {\color{darkgreen} \underline{\textbf{77.22}}} & 68.72 & 65.98 & 60.76 & 60.49 & 56.00 & 91.98 & 58.81 \\
& \texttt{MULTIVERS}\small{~\cite{DBLP:conf/naacl/WaddenLWCBH22}} & 68.86 & 60.17 & 59.87 & 52.55 & 55.67 & 51.86 & 86.03 & 56.61 \\
 \midrule
\multirow{2}{*}{\uppercase\expandafter{\romannumeral3}} & \texttt{Codex}\small{~\cite{DBLP:journals/corr/abs-2107-03374}} & 70.63 & 65.07 & 66.46 & 56.63 & 63.49 & 57.27 & 89.77 & 62.58 \\
& \texttt{FLAN-T5}\small{~\cite{DBLP:journals/corr/abs-2210-11416}} & 73.69 & 69.02 & 65.66 & 60.23 & 58.08 & 55.42 & 90.81 & 63.73 \\
 \midrule
\multirow{2}{*}{\uppercase\expandafter{\romannumeral4}} & \texttt{ProgramFC (N=1)} & 74.10 & 69.36 & 66.13 & 60.63 & 65.69 & {\color{darkgreen} \underline{\textbf{59.16}}} & 91.77 & 67.80 \\
& \texttt{ProgramFC (N=5)} 
    & 75.65 
    & {\color{darkgreen} \underline{\textbf{70.30}}} 
    & {\color{darkgreen} \underline{\textbf{68.48}}} 
    & {\color{darkgreen} \underline{\textbf{63.43}}} 
    & {\color{darkgreen} \underline{\textbf{66.75}}} 
    & 57.74 
    & {\color{darkgreen} \underline{\textbf{92.69}}} 
    & {\color{darkgreen} \underline{\textbf{68.06}}} \\ \bottomrule
\end{tabular}%
}
\caption{Macro-F1 scores of \model (\uppercase\expandafter{\romannumeral4}) and baselines (\uppercase\expandafter{\romannumeral1}-\uppercase\expandafter{\romannumeral3}) on the evaluation set of HOVER and FEVEROUS-S for few-shot fact-checking. \textit{Gold} and \textit{Open} represent the gold evidence setting and the open book setting, respectively. \uppercase\expandafter{\romannumeral1}: pretrained Transformers; \uppercase\expandafter{\romannumeral2}: FC/NLI fine-tuned models; \uppercase\expandafter{\romannumeral3}: in-context learning models. }  
\label{tbl:main_results}
\end{table*}

\subsection{Main Results}
\label{sec:main_results}

We report the overall results for \model and for the baselines for few-shot fact-checking in Table~\ref{tbl:main_results}. \model achieves the best performance on 7 out of 8 evaluations, demonstrating its effectiveness. We have three more specific observations. 

\vspace{0.1cm}

\noindent \underline{\textbf{\texttt{ProgramFC} is more effective on deeper claims.}}

\noindent On the HOVER dataset, \texttt{ProgramFC (N=5)} outperforms the baselines on average by 10.38\%, 11.37\%, and 14.77\% on two-hop, three-hop, and four-hop claims, respectively. 
This suggests that \texttt{ProgramFC} becomes increasingly effective as the required reasoning depth increases. Among the baselines, \texttt{DeBERTaV3-NLI} performs comparably to \texttt{ProgramFC} on two-hop claims, indicating that large-scale pre-training on simpler claims can help the model generalize to more complex claims. 

However, this generalization becomes more challenging as the complexity of the claims increases. On HOVER, the F1 score of \texttt{DeBERTaV3-NLI} drops from 77.22 for 2-hop claims to 60.49 for 4-hop claims, which is a decrease of 21.7\%. In contrast, the performance drop for \texttt{ProgramFC}, which uses the strategy of program-guided reasoning, is much smaller: just 11.7\%. 


\vspace{0.1cm}

\noindent \ul{\textbf{Decomposition is more effective than one-step prediction.}} 
The \texttt{ProgramFC} model, which uses the same FLAN-T5 model as the sub-task functions, outperforms the baseline of directly verifying claims with \texttt{FLAN-T5} on all four datasets. On average, there is a 6.0\% improvement in the gold evidence setting and a 4.5\% improvement in the open-book setting. This suggests that decomposing a complex claim into simpler steps with a program can facilitate more accurate reasoning. This is especially evident when the required reasoning is complex: there is a 14.9\% improvement in the gold evidence setting and a 6.7\% improvement in the open-book setting for 4-hop claims. 

\vspace{0.1cm}

\noindent \underline{\textbf{Aggregating reasoning programs is helpful.}}


\noindent We find that aggregating the predictions of $N=5$ reasoning programs improves the performance over using a single program by an average of 1.5\%. This aligns with the findings of~\citet{DBLP:journals/corr/abs-2203-11171}, where the idea was applied for question answering: if multiple different ways of thinking lead to the same answer, we can have greater confidence that the final answer is correct. This intuition also applies to fact-checking, as each program represents a unique reasoning chain to verify the claim. 


\begin{figure*}[!t]
	\centering
	\includegraphics[width=16cm]{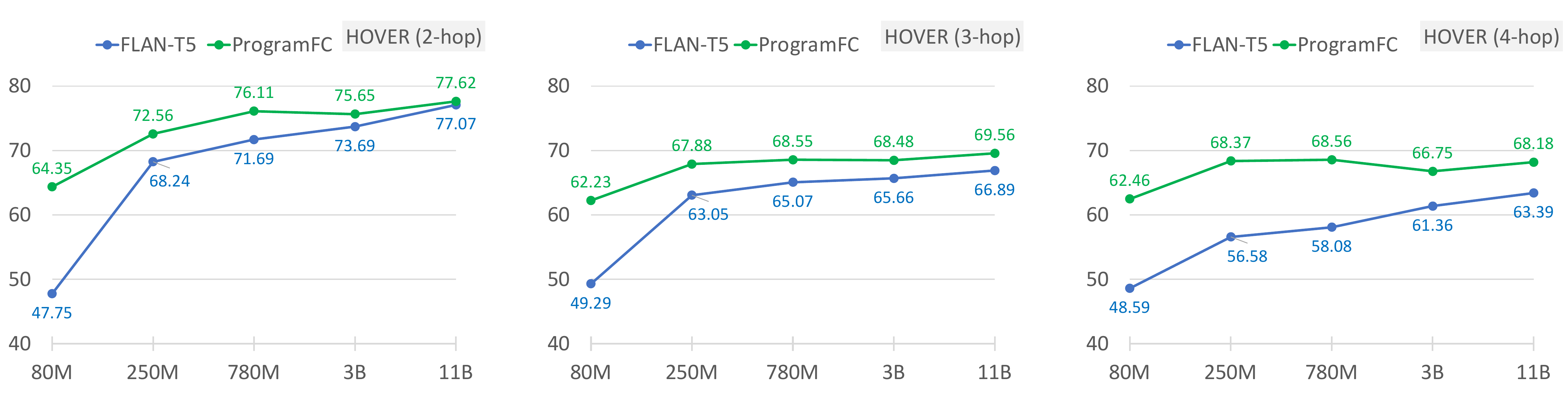}
    \caption{F1 score for fact-checking with gold evidence using \texttt{FLAN-T5} (blue line) and \model (green line) for language models of increasing sizes: \texttt{FLAN-T5-small} (80M), \texttt{FLAN-T5-base} (250M), \texttt{FLAN-large} (780M), \texttt{FLAN-T5-XL} (3B), and \texttt{FLAN-T5-XXL} (11B) on HOVER 2-hop (left), 3-hop (middle), and 4-hop (right).}
    \label{fig:model_size}
\end{figure*}

\begin{figure}[!t]
	\centering
	\includegraphics[width=7.5cm]{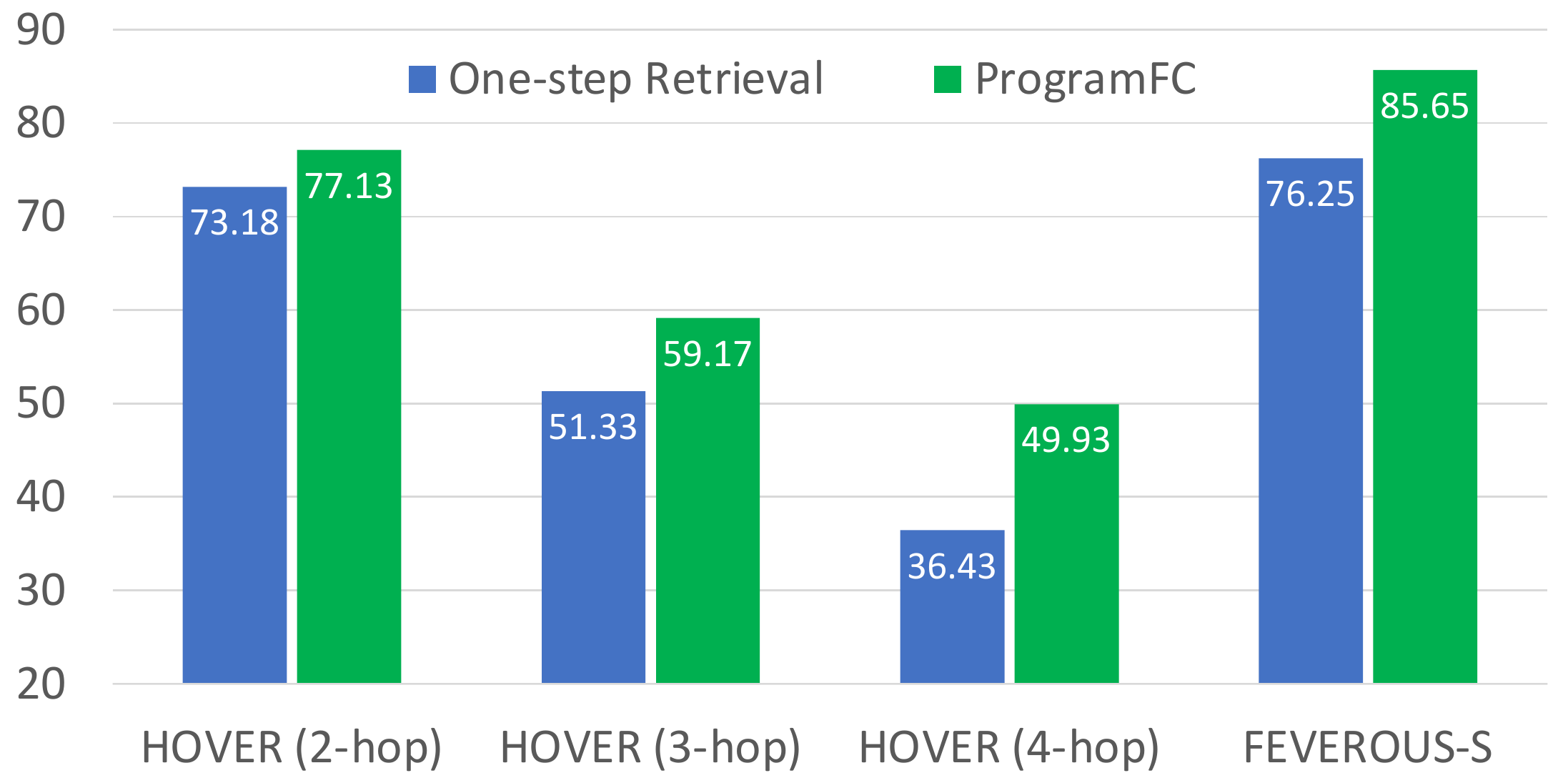}
    \caption{Retrieval recall@10 for the one-step retrieval and the iterative retrieval in \model.}
    \label{fig:retriever_performance}
    \vspace{-0.2cm}
\end{figure}

\subsection{How Does the Reasoning Program Help?}
\label{sec:how_reasoning_program_help}

To further understand how reasoning programs facilitate fact-checking, we compare the performance of \model with \texttt{FLAN-T5} using different language model sizes: \texttt{small}, \texttt{base}, \texttt{large}, \texttt{XL}, and \texttt{XXL}. The results are shown in Figure~\ref{fig:model_size} and indicate that program-guided reasoning is particularly effective when the model size is small. As smaller models have less capacity for complex reasoning, the performance of the end-to-end \texttt{FLAN-T5} model decreases significantly with decreasing model size. However, this trend is less notable for \model. The high-level reasoning plan offered by reasoning programs substantially alleviates the demands on the subsequent sub-task solvers. Our results show that the program-guided model using \texttt{FLAN-T5-small} (80M parameters) as sub-task solvers can achieve comparable performance to the 137x larger \texttt{FLAN-T5-XXL} (11B) model with end-to-end reasoning for 4-hop claims. 



In the open-domain setting, we find that reasoning programs can enhance the retrieval of relevant evidence from the knowledge source. Figure~\ref{fig:retriever_performance} compares the retrieval performance of the one-step BM25 retriever used in the baselines to the iterative step-by-step BM25 retriever in \model. 

We measure the recall of the gold paragraphs for the top-10 retrieved paragraphs (recall@10). For \model, we combine the retrieved paragraphs of all steps and we consider the top-10 results. We can see in Figure~\ref{fig:retriever_performance} that \model outperforms one-step retrieval on all datasets, with the largest improvement of 37.1\% on HOVER 4-hop. This is because some information may not be present in the original claim, but is only revealed during the reasoning process (\textit{e.g.},~``Christopher Nolan'' in Figure~\ref{fig:general_framework}). Thus, iterative retrieval guided by the reasoning program yields better results. 

\subsection{Interpretability of Reasoning Programs}
\label{sec:error_analysis}

An advantage of \model is that it improves the interpretability of fact-checking compared to end-to-end models, as the explicit program can aid human understanding and debugging. Examples of generated reasoning programs can be found in Figure~\ref{fig:correct_examples} of Appendix~\ref{appendix:correct_examples}. To assess the quality of the generated reasoning programs, we sampled 300 claims where \model \textit{incorrectly} predicted the final veracity labels from the HOVER 2-hop, 3-hop, and 4-hop datasets, with 100 examples per dataset. We asked human annotators to analyze the error types and we classified the results into three categories: (\emph{i})~\textit{Syntactic errors}, where the program does not conform to the defined grammar and cannot be parsed, (\emph{ii})~\textit{Semantic errors}, which include incorrect or missing arguments/variables (\textit{Token}), incorrect program structure (\textit{Structure}), and incorrect sub-task calls (\textit{Subtask}), and (\emph{iii})~\textit{Incorrect execution}, where the program is correct, but where the incorrect prediction is a result of its execution. 

\begin{table}[!t]
\centering
\resizebox{0.48\textwidth}{!}{
    \begin{tabular}{l|c|c|l}
    \toprule
        \multirow{2}{*}{\textbf{Error Type}} & \multicolumn{3}{c}{\textbf{Proportion (\%)}} \\ \cline{2-4}
         & 2-hop & 3-hop & 4-hop \\ \midrule
         Syntax error & \textbf{0\%} & \textbf{0\%} & \textbf{0\%} \\
         Semantic error & \textbf{29\%} & \textbf{38\%} & \textbf{77\%} \\
         \quad \quad Token  & \quad \quad 8\% & \quad \quad 20\% & \quad \quad 18\% \\
         \quad \quad Structure & \quad \quad 19\% & \quad \quad 13\% & \quad \quad 57\% \\
         \quad \quad Subtask & \quad \quad 2\% & \quad \quad 5\% & \quad \quad 2\% \\
         Incorrect execution & \textbf{71\%} & \textbf{62\%} & \textbf{23\%} \\ \bottomrule
    \end{tabular}
    }
    \caption{Reasoning program evaluation for incorrectly-predicted examples from each hop length in HOVER.}
    \label{tbl:error_analysis}
\end{table}

We show the error analysis in Table~\ref{tbl:error_analysis}. First, no syntax errors were found in our samples, indicating that Codex effectively generates executable programs through few-shot in-context learning. 

Second, for 2-hop claims, 71\% of the programs are found to be correct and the majority of the errors are the result of incorrect program execution, where the question answering or the fact-checking modules failed to return the correct answer. 

\begin{figure*}[!t]
	\centering
	\includegraphics[width=15cm]{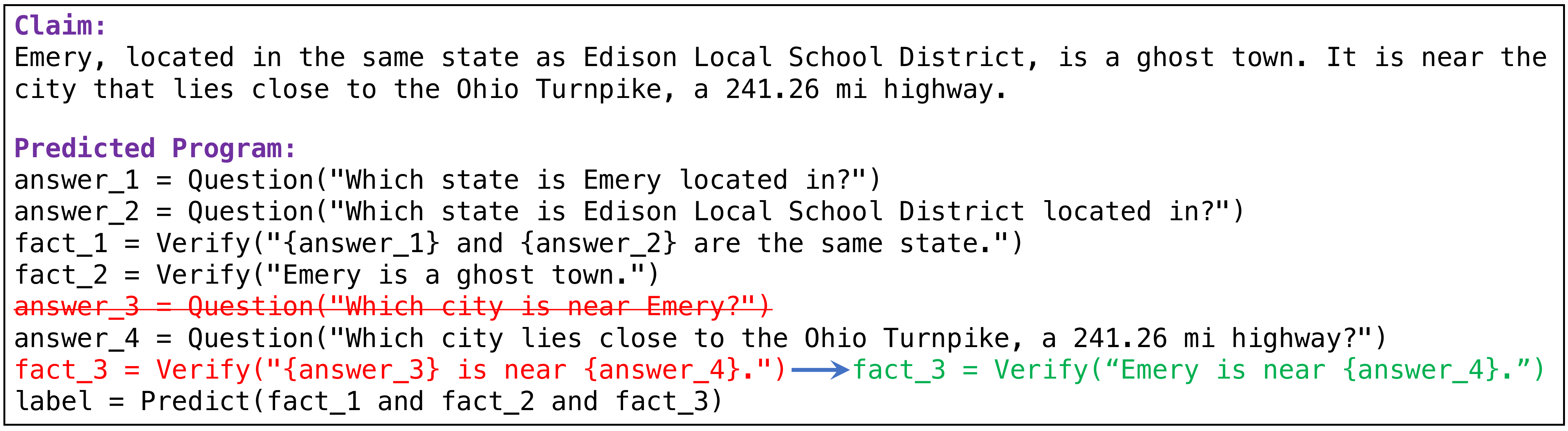}
    \caption{An error case from the HOVER 4-hop dataset where the generated reasoning program has an incorrect program structure. The incorrect segment(s) are marked in {\color{red} \bf red}, and the correct revisions are marked in {\color{darkgreen} \bf green}.}
    \label{fig:case_study}
\end{figure*}

Third, as the complexity of the claims increased, the proportion of semantic errors in the programs increased, with structural errors becoming particularly prevalent. This highlights the difficulty of generating the appropriate step-by-step reasoning strategies for claims that require long-chain reasoning. An example structural error is shown in Figure~\ref{fig:case_study}, where the model fails to parse the second sentence of the claim into correct program instructions. Additional error examples can be found in Appendix~\ref{appendix:examples_programs}. 

\subsection{Closed-Book Fact-Checking}
\label{sec:close_book}

Finally, we evaluate the closed-book setting, where the model does not have access to any knowledge source and needs to rely on its parametric knowledge only. The baseline models from groups \uppercase\expandafter{\romannumeral1} and \uppercase\expandafter{\romannumeral2} in Table~\ref{tbl:main_results} are trained with (evidence, claim) pairs and thus are not applicable in this setting. We compare our method to the baselines that use large language models for in-context learning, including \texttt{Codex} (\texttt{code-davinci-002}) and \texttt{FLAN-T5} from Table~\ref{tbl:main_results}. 

We also include the 175B-parameter InstructGPT (\texttt{text-davinci-002})~\cite{DBLP:journals/corr/abs-2203-02155} with four different prompts: (\emph{i})~\textit{direct} prompting with the claim, (\emph{ii})~\texttt{CoT}~\cite{DBLP:journals/corr/abs-2201-11903} or chain-of-thought prompting with demonstrations, (\emph{iii})~\texttt{ZS-CoT}~\cite{DBLP:journals/corr/abs-2205-11916} or zero-shot chain-of-thought with the prompt ``let's think step by step'', and (\emph{iv})~\texttt{Self-Ask}~\cite{DBLP:journals/corr/abs-2210-03350}, which is a variant of CoT that guides the model reasoning by asking a series of questions. The detailed prompting templates are given in Appendix~\ref{appendix:closed_book_prompt}. 

\begin{table}[!t]
\centering
\resizebox{0.48\textwidth}{!}{
\renewcommand{\arraystretch}{1.1}
\begin{tabular}{l|cccc}
\toprule
\multirow{2}{*}{Model} & \multicolumn{3}{c}{HOVER} & \multirow{2}{*}{FEVEROUS} \\
 & 2-hop & 3-hop & 4-hop & \\ \midrule
\texttt{InstructGPT} & & & & \\
\quad \texttt{- Direct} & 56.51 & 51.75 & 49.68 & 60.13 \\
\quad \texttt{- ZS-CoT} & 50.30 & 52.30 & 51.58 & 54.78 \\
\quad \texttt{- CoT} & {\color{darkgreen} \underline{\textbf{57.20}}} & 53.66 & 51.83 & {\color{darkgreen} \underline{\textbf{61.05}}} \\ 
\quad \texttt{- Self-Ask} & 51.54 & 51.47 & 52.45 & 56.82 \\ 
\texttt{Codex} & 55.57 & 53.42 & 45.59 & 57.85 \\
\texttt{FLAN-T5} & 48.27 & 52.11 & 51.13 & 55.16 \\ \midrule
\texttt{ProgramFC} & 54.27 & {\color{darkgreen} \underline{\textbf{54.18}}} & {\color{darkgreen} \underline{\textbf{52.88}}} & 59.66 \\ \bottomrule
\end{tabular}%
}
\caption{Closed-book setting: macro-F1 scores for \model and for the baselines.}
\label{tbl:closed_book_results}
\vspace{-0.1cm}
\end{table}

Our results, presented in Table~\ref{tbl:closed_book_results}, show that most models achieve a Macro-F1 score only slightly above random guessing on the HOVER dataset, indicating the difficulty of solely relying on parametric knowledge of large language models for fact-checking complex claims. Similarly to the observations in Section~\ref{sec:main_results}, we see a trend of improved performance as the number of the required reasoning hops increases. Chain-of-thought prompting scores an average 2.7 points higher than direct prompting, highlighting the importance of step-by-step reasoning for complex fact-checking. It outperforms our \model on HOVER 2-hop and FEVEROUS but performs worse on HOVER 3-hop and 4-hop. 

This can be due to CoT generating free-form explanations, which can lead to unpredictable errors in long reasoning chains. In contrast, our program generation-and-execution strategy is more stable for longer reasoning chains. 


\section{Conclusion and Future Work}

We proposed \model, a few-shot neuro-symbolic model for fact-checking that learns to map input claims to a reasoning program consisting of a sequence of sub-task function calls for answering a question, for fact-checking a simple claim, and for computing a logical expression. Then fact-checking is performed by executing that program. \model combines the advantages of symbolic programs, such as explainability, with the flexibility of end-to-end neural models. Using Codex as the program generator, \model demonstrates promising performance on HOVER and FEVEROUS with only a small number of in-context demonstrations and no additional training. We also investigated the impact of model size and the benefits of programs for retrieval, and we analyzed the errors. The results indicated that \model effectively balances model capability, learning efficiency, and interpretability.

In future work, we want to adapt \model to more real-world fact-checking scenarios, such as fake news detection and multi-modal fact-checking, with advanced reasoning program design and sub-task functionalities.

\section*{Limitations}
We identify two main limitations of \model. First, the claims in the HOVER and the FEVEROUS datasets, despite being complex in their surface form, mostly only require \textit{explicit} multi-step reasoning, \textit{i.e.},~the decomposition can be derived from the claim's syntactic structure or how the claim is framed. This lowers the difficulty of generating reasoning programs. However, for many real-world complex claims, the reasoning is often \textit{implicit}. For example, for the claim \emph{``Aristotle couldn't have used a laptop''}, the reasoning program looks as follows:

\noindent answer\_1 = Question(``When did Aristotle live?'');

\vspace{0.1cm}
\noindent answer\_2 = Question(``When was the laptop invented?'');

\vspace{0.1cm}
\noindent fact\_1 = Verify(``{answer\_1} is before {answer\_2}.'');

\vspace{0.1cm}
\noindent label = Predict(fact\_1)
\\
Generating reasoning programs for such implicit complex claims requires a deeper understanding of the claim and also access to world and commonsense knowledge. We conducted preliminary experiments on these types of claims, but we found that our Codex-based generator struggled to produce a correct reasoning program. This highlights the gap in applying our \model to fact-check real-world claims. Addressing these challenges is an important direction for future work. 

Second, we have found that \model has a higher computational cost than the baseline end-to-end fact-checking models. It requires calling large language models for program generation and further calling multiple sub-task models. This results in the actual computational time that is $\sim$4-5$\times$ higher than for an end-to-end \texttt{FLAN-T5} model. Thus, developing more efficient methods for program generation and execution is one of the important directions for future work. 

\section*{Ethics Statement}

\paragraph{Biases.}

We note that there might be some biases in the data used to train the LLMs, as well as in some factuality judgments. Both of these are beyond our control.

\paragraph{Intended Use and Misuse Potential.}

Our models can be of interest to the general public and could also save a lot of time to human fact-checkers. However, they could also be misused by malicious actors. We, therefore, ask researchers to exercise caution.

\paragraph{Environmental Impact.}
We would like to warn that the use of large language models requires a significant amount of computation and the use of GPUs/TPUs for training, which contributes to global warming. 
This is a bit less of an issue in our case, as we do not train such models from scratch; rather, we perform few-shot in-context learning. Still, the large language model (Codex) we are calling is likely running on a GPU.

\section*{Acknowledgements}
This work was supported by the National Science Foundation award \#2048122. The views expressed are those of the author and do not reflect the official policy or position of the US government. We thank Alex Mei, Xinyi Wang, Danqing Wang, Sharon Levy, Gyuwan Kim, and other members of the UCSB NLP group for their valuable feedback. 

\bibliography{anthology,custom}

\begin{thebibliography}{70}
\expandafter\ifx\csname natexlab\endcsname\relax\def\natexlab#1{#1}\fi

\bibitem[{Ahmadi et~al.(2019)Ahmadi, Lee, Papotti, and
  Saeed}]{DBLP:conf/tto/AhmadiLPS19}
Naser Ahmadi, Joohyung Lee, Paolo Papotti, and Mohammed Saeed. 2019.
\newblock \href
  {https://truthandtrustonline.com/wp-content/uploads/2019/09/paper\_15.pdf}
  {Explainable fact checking with probabilistic answer set programming}.
\newblock In \emph{Proceedings of the Truth and Trust Online Conference (TTO)},
  London, UK.

\bibitem[{Aly et~al.(2021)Aly, Guo, Schlichtkrull, Thorne, Vlachos,
  Christodoulopoulos, Cocarascu, and Mittal}]{DBLP:conf/nips/AlyGST00CM21}
Rami Aly, Zhijiang Guo, Michael~Sejr Schlichtkrull, James Thorne, Andreas
  Vlachos, Christos Christodoulopoulos, Oana Cocarascu, and Arpit Mittal. 2021.
\newblock \href
  {https://datasets-benchmarks-proceedings.neurips.cc/paper/2021/hash/68d30a9594728bc39aa24be94b319d21-Abstract-round1.html}
  {{FEVEROUS:} {Fact Extraction and VERification Over Unstructured and
  Structured information}}.
\newblock In \emph{Proceedings of the Neural Information Processing Systems
  (NeurIPS) Track on Datasets and Benchmarks}, Online.

\bibitem[{Aly and Vlachos(2022)}]{aly-vlachos-2022-natural}
Rami Aly and Andreas Vlachos. 2022.
\newblock \href {https://aclanthology.org/2022.emnlp-main.411} {Natural
  logic-guided autoregressive multi-hop document retrieval for fact
  verification}.
\newblock In \emph{Proceedings of the 2022 Conference on Empirical Methods in
  Natural Language Processing (EMNLP)}, pages 6123--6135, Abu Dhabi, United
  Arab Emirates.

\bibitem[{Atanasova et~al.(2020)Atanasova, Simonsen, Lioma, and
  Augenstein}]{DBLP:conf/acl/AtanasovaSLA20}
Pepa Atanasova, Jakob~Grue Simonsen, Christina Lioma, and Isabelle Augenstein.
  2020.
\newblock \href {https://doi.org/10.18653/v1/2020.acl-main.656} {Generating
  fact checking explanations}.
\newblock In \emph{Proceedings of the 58th Annual Meeting of the Association
  for Computational Linguistics (ACL)}, pages 7352--7364, Online.

\bibitem[{Augenstein et~al.(2019)Augenstein, Lioma, Wang, Chaves~Lima, Hansen,
  Hansen, and Simonsen}]{augenstein-etal-2019multifc}
Isabelle Augenstein, Christina Lioma, Dongsheng Wang, Lucas Chaves~Lima, Casper
  Hansen, Christian Hansen, and Jakob~Grue Simonsen. 2019.
\newblock \href {https://doi.org/10.18653/v1/D19-1475} {{M}ulti{FC}: A
  real-world multi-domain dataset for evidence-based fact checking of claims}.
\newblock In \emph{Proceedings of the 2019 Conference on Empirical Methods in
  Natural Language Processing and the 9th International Joint Conference on
  Natural Language Processing (EMNLP-IJCNLP)}, pages 4685--4697, Hong Kong,
  China.

\bibitem[{Barnabò et~al.(2022)Barnabò, Siciliano, Castillo, Leonardi, Nakov,
  Da~San~Martino, and Silvestri}]{FbMultiLingMisinfo:2022}
Giorgio Barnabò, Federico Siciliano, Carlos Castillo, Stefano Leonardi,
  Preslav Nakov, Giovanni Da~San~Martino, and Fabrizio Silvestri. 2022.
\newblock \href {https://doi.org/10.1109/IJCNN55064.2022.9892739}
  {{FbMultiLingMisinfo}: Challenging large-scale multilingual benchmark for
  misinformation detection}.
\newblock In \emph{Proceedings of the 2022 International Joint Conference on
  Neural Networks (IJCNN)}, pages 1--8, Padova, Italy.

\bibitem[{Barnabò et~al.(2023)Barnabò, Siciliano, Castillo, Leonardi, Nakov,
  {Da San Martino}, and Silvestri}]{BARNABO2023100244}
Giorgio Barnabò, Federico Siciliano, Carlos Castillo, Stefano Leonardi,
  Preslav Nakov, Giovanni {Da San Martino}, and Fabrizio Silvestri. 2023.
\newblock \href {https://doi.org/https://doi.org/10.1016/j.osnem.2023.100244}
  {Deep active learning for misinformation detection using geometric deep
  learning}.
\newblock \emph{Online Social Networks and Media}, 33:100244.

\bibitem[{Beltagy et~al.(2020)Beltagy, Peters, and
  Cohan}]{DBLP:journals/corr/abs-2004-05150}
Iz~Beltagy, Matthew~E. Peters, and Arman Cohan. 2020.
\newblock \href {https://arxiv.org/abs/2004.05150} {Longformer: The
  long-document transformer}.
\newblock \emph{ArXiv preprint}, abs/2004.05150.

\bibitem[{Bowman et~al.(2015)Bowman, Angeli, Potts, and
  Manning}]{DBLP:conf/emnlp/BowmanAPM15}
Samuel~R. Bowman, Gabor Angeli, Christopher Potts, and Christopher~D. Manning.
  2015.
\newblock \href {https://doi.org/10.18653/v1/D15-1075} {A large annotated
  corpus for learning natural language inference}.
\newblock In \emph{Proceedings of the 2015 Conference on Empirical Methods in
  Natural Language Processing (EMNLP)}, pages 632--642, Lisbon, Portugal.

\bibitem[{Brown et~al.(2020)Brown, Mann, Ryder, Subbiah, Kaplan, Dhariwal,
  Neelakantan, Shyam, Sastry, Askell, Agarwal, Herbert{-}Voss, Krueger,
  Henighan, Child, Ramesh, Ziegler, Wu, Winter, Hesse, Chen, Sigler, Litwin,
  Gray, Chess, Clark, Berner, McCandlish, Radford, Sutskever, and
  Amodei}]{DBLP:conf/nips/BrownMRSKDNSSAA20}
Tom~B. Brown, Benjamin Mann, Nick Ryder, Melanie Subbiah, Jared Kaplan,
  Prafulla Dhariwal, Arvind Neelakantan, Pranav Shyam, Girish Sastry, Amanda
  Askell, Sandhini Agarwal, Ariel Herbert{-}Voss, Gretchen Krueger, Tom
  Henighan, Rewon Child, Aditya Ramesh, Daniel~M. Ziegler, Jeffrey Wu, Clemens
  Winter, Christopher Hesse, Mark Chen, Eric Sigler, Mateusz Litwin, Scott
  Gray, Benjamin Chess, Jack Clark, Christopher Berner, Sam McCandlish, Alec
  Radford, Ilya Sutskever, and Dario Amodei. 2020.
\newblock \href
  {https://proceedings.neurips.cc/paper/2020/hash/1457c0d6bfcb4967418bfb8ac142f64a-Abstract.html}
  {Language models are few-shot learners}.
\newblock In \emph{Proceedings of the Annual Conference on Neural Information
  Processing Systems (NeurIPS)}, Online.

\bibitem[{Chen et~al.(2022{\natexlab{a}})Chen, Sriram, Choi, and
  Durrett}]{Chen2022GeneratingLA}
Jifan Chen, Aniruddh Sriram, Eunsol Choi, and Greg Durrett. 2022{\natexlab{a}}.
\newblock \href {https://aclanthology.org/2022.emnlp-main.229} {Generating
  literal and implied subquestions to fact-check complex claims}.
\newblock In \emph{Proceedings of the 2022 Conference on Empirical Methods in
  Natural Language Processing (EMNLP)}, pages 3495--3516, Abu Dhabi, United
  Arab Emirates.

\bibitem[{Chen et~al.(2021)Chen, Tworek, Jun, Yuan, de~Oliveira~Pinto, Kaplan,
  Edwards, Burda, Joseph, Brockman, Ray, Puri, Krueger, Petrov, Khlaaf, Sastry,
  Mishkin, Chan, Gray, Ryder, Pavlov, Power, Kaiser, Bavarian, Winter, Tillet,
  Such, Cummings, Plappert, Chantzis, Barnes, Herbert{-}Voss, Guss, Nichol,
  Paino, Tezak, Tang, Babuschkin, Balaji, Jain, Saunders, Hesse, Carr, Leike,
  Achiam, Misra, Morikawa, Radford, Knight, Brundage, Murati, Mayer, Welinder,
  McGrew, Amodei, McCandlish, Sutskever, and
  Zaremba}]{DBLP:journals/corr/abs-2107-03374}
Mark Chen, Jerry Tworek, Heewoo Jun, Qiming Yuan, Henrique~Ponde
  de~Oliveira~Pinto, Jared Kaplan, Harrison Edwards, Yuri Burda, Nicholas
  Joseph, Greg Brockman, Alex Ray, Raul Puri, Gretchen Krueger, Michael Petrov,
  Heidy Khlaaf, Girish Sastry, Pamela Mishkin, Brooke Chan, Scott Gray, Nick
  Ryder, Mikhail Pavlov, Alethea Power, Lukasz Kaiser, Mohammad Bavarian,
  Clemens Winter, Philippe Tillet, Felipe~Petroski Such, Dave Cummings,
  Matthias Plappert, Fotios Chantzis, Elizabeth Barnes, Ariel Herbert{-}Voss,
  William~Hebgen Guss, Alex Nichol, Alex Paino, Nikolas Tezak, Jie Tang, Igor
  Babuschkin, Suchir Balaji, Shantanu Jain, William Saunders, Christopher
  Hesse, Andrew~N. Carr, Jan Leike, Joshua Achiam, Vedant Misra, Evan Morikawa,
  Alec Radford, Matthew Knight, Miles Brundage, Mira Murati, Katie Mayer, Peter
  Welinder, Bob McGrew, Dario Amodei, Sam McCandlish, Ilya Sutskever, and
  Wojciech Zaremba. 2021.
\newblock \href {https://arxiv.org/abs/2107.03374} {Evaluating large language
  models trained on code}.
\newblock \emph{ArXiv preprint}, abs/2107.03374.

\bibitem[{Chen et~al.(2022{\natexlab{b}})Chen, Ma, Wang, and
  Cohen}]{DBLP:journals/corr/abs-2211-12588}
Wenhu Chen, Xueguang Ma, Xinyi Wang, and William~W. Cohen. 2022{\natexlab{b}}.
\newblock \href {https://doi.org/10.48550/arXiv.2211.12588} {Program of
  thoughts prompting: Disentangling computation from reasoning for numerical
  reasoning tasks}.
\newblock \emph{CoRR}, abs/2211.12588.

\bibitem[{Cheng et~al.(2022)Cheng, Xie, Shi, Li, Nadkarni, Hu, Xiong, Radev,
  Ostendorf, Zettlemoyer, Smith, and Yu}]{DBLP:journals/corr/abs-2210-02875}
Zhoujun Cheng, Tianbao Xie, Peng Shi, Chengzu Li, Rahul Nadkarni, Yushi Hu,
  Caiming Xiong, Dragomir Radev, Mari Ostendorf, Luke Zettlemoyer, Noah~A.
  Smith, and Tao Yu. 2022.
\newblock \href {https://doi.org/10.48550/arXiv.2210.02875} {Binding language
  models in symbolic languages}.
\newblock \emph{CoRR}, abs/2210.02875.

\bibitem[{Chung et~al.(2022)Chung, Hou, Longpre, Zoph, Tay, Fedus, Li, Wang,
  Dehghani, Brahma, Webson, Gu, Dai, Suzgun, Chen, Chowdhery, Narang, Mishra,
  Yu, Zhao, Huang, Dai, Yu, Petrov, Chi, Dean, Devlin, Roberts, Zhou, Le, and
  Wei}]{DBLP:journals/corr/abs-2210-11416}
Hyung~Won Chung, Le~Hou, Shayne Longpre, Barret Zoph, Yi~Tay, William Fedus,
  Eric Li, Xuezhi Wang, Mostafa Dehghani, Siddhartha Brahma, Albert Webson,
  Shixiang~Shane Gu, Zhuyun Dai, Mirac Suzgun, Xinyun Chen, Aakanksha
  Chowdhery, Sharan Narang, Gaurav Mishra, Adams Yu, Vincent~Y. Zhao, Yanping
  Huang, Andrew~M. Dai, Hongkun Yu, Slav Petrov, Ed~H. Chi, Jeff Dean, Jacob
  Devlin, Adam Roberts, Denny Zhou, Quoc~V. Le, and Jason Wei. 2022.
\newblock \href {https://doi.org/10.48550/arXiv.2210.11416} {Scaling
  instruction-finetuned language models}.
\newblock \emph{CoRR}, abs/2210.11416.

\bibitem[{Cui et~al.(2019)Cui, Shu, Wang, Lee, and
  Liu}]{DBLP:conf/cikm/CuiSW0L19}
Limeng Cui, Kai Shu, Suhang Wang, Dongwon Lee, and Huan Liu. 2019.
\newblock \href {https://doi.org/10.1145/3357384.3357862} {{dEFEND}: {A} system
  for explainable fake news detection}.
\newblock In \emph{Proceedings of the 28th {ACM} International Conference on
  Information and Knowledge Management (CIKM)}, pages 2961--2964, Beijing,
  China.

\bibitem[{Devlin et~al.(2019)Devlin, Chang, Lee, and
  Toutanova}]{DBLP:conf/naacl/DevlinCLT19}
Jacob Devlin, Ming-Wei Chang, Kenton Lee, and Kristina Toutanova. 2019.
\newblock \href {https://doi.org/10.18653/v1/N19-1423} {{BERT}: Pre-training of
  deep bidirectional transformers for language understanding}.
\newblock In \emph{Proceedings of the 2019 Conference of the North {A}merican
  Chapter of the Association for Computational Linguistics: Human Language
  Technologies (NAACL-HLT)}, pages 4171--4186, Minneapolis, Minnesota, USA.

\bibitem[{Gad{-}Elrab et~al.(2019)Gad{-}Elrab, Stepanova, Urbani, and
  Weikum}]{DBLP:conf/wsdm/Gad-Elrab0UW19}
Mohamed~H. Gad{-}Elrab, Daria Stepanova, Jacopo Urbani, and Gerhard Weikum.
  2019.
\newblock \href {https://doi.org/10.1145/3289600.3290996} {Exfakt: {A}
  framework for explaining facts over knowledge graphs and text}.
\newblock In \emph{Proceedings of the Twelfth {ACM} International Conference on
  Web Search and Data Mining (WSDM)}, pages 87--95, Melbourne, Australia.

\bibitem[{Gao et~al.(2022)Gao, Madaan, Zhou, Alon, Liu, Yang, Callan, and
  Neubig}]{DBLP:journals/corr/abs-2211-10435}
Luyu Gao, Aman Madaan, Shuyan Zhou, Uri Alon, Pengfei Liu, Yiming Yang, Jamie
  Callan, and Graham Neubig. 2022.
\newblock {PAL:} program-aided language models.
\newblock \emph{CoRR}, abs/2211.10435.

\bibitem[{Glockner et~al.(2022)Glockner, Hou, and
  Gurevych}]{glockner-etal-2022-missing}
Max Glockner, Yufang Hou, and Iryna Gurevych. 2022.
\newblock \href {https://aclanthology.org/2022.emnlp-main.397} {Missing
  counter-evidence renders {NLP} fact-checking unrealistic for misinformation}.
\newblock In \emph{Proceedings of the 2022 Conference on Empirical Methods in
  Natural Language Processing (EMNLP)}, pages 5916--5936, Abu Dhabi, United
  Arab Emirates.

\bibitem[{Guo et~al.(2022)Guo, Schlichtkrull, and
  Vlachos}]{DBLP:journals/tacl/GuoSV22}
Zhijiang Guo, Michael Schlichtkrull, and Andreas Vlachos. 2022.
\newblock \href {https://doi.org/10.1162/tacl_a_00454} {A survey on automated
  fact-checking}.
\newblock \emph{Transactions of the Association for Computational Linguistics},
  10:178--206.

\bibitem[{Gupta and Srikumar(2021)}]{DBLP:conf/acl/GuptaS20}
Ashim Gupta and Vivek Srikumar. 2021.
\newblock \href {https://doi.org/10.18653/v1/2021.acl-short.86} {{X}-{F}act: A
  new benchmark dataset for multilingual fact checking}.
\newblock In \emph{Proceedings of the 59th Annual Meeting of the Association
  for Computational Linguistics and the 11th International Joint Conference on
  Natural Language Processing (ACL-IJCNLP)}, pages 675--682, Online.

\bibitem[{He et~al.(2021)He, Gao, and Chen}]{DBLP:journals/corr/abs-2111-09543}
Pengcheng He, Jianfeng Gao, and Weizhu Chen. 2021.
\newblock \href {https://arxiv.org/abs/2111.09543} {{DeBERTaV3}: Improving
  {DeBERTa} using {ELECTRA}-style pre-training with gradient-disentangled
  embedding sharing}.
\newblock \emph{ArXiv preprint}, abs/2111.09543.

\bibitem[{Jiang et~al.(2021)Jiang, Pradeep, and Lin}]{DBLP:conf/acl/JiangPL20}
Kelvin Jiang, Ronak Pradeep, and Jimmy Lin. 2021.
\newblock \href {https://doi.org/10.18653/v1/2021.acl-short.51} {Exploring
  listwise evidence reasoning with {T5} for fact verification}.
\newblock In \emph{Proceedings of the 59th Annual Meeting of the Association
  for Computational Linguistics and the 11th International Joint Conference on
  Natural Language Processing (ACL-IJCNLP)}, pages 402--410, Online.

\bibitem[{Jiang et~al.(2020)Jiang, Bordia, Zhong, Dognin, Singh, and
  Bansal}]{Jiang2020HoVerAD}
Yichen Jiang, Shikha Bordia, Zheng Zhong, Charles Dognin, Maneesh Singh, and
  Mohit Bansal. 2020.
\newblock \href {https://doi.org/10.18653/v1/2020.findings-emnlp.309}
  {{H}o{V}er: A dataset for many-hop fact extraction and claim verification}.
\newblock In \emph{Findings of the Association for Computational Linguistics:
  EMNLP 2020}, pages 3441--3460, Online.

\bibitem[{Jolly et~al.(2022)Jolly, Atanasova, and
  Augenstein}]{DBLP:journals/information/JollyAA22}
Shailza Jolly, Pepa Atanasova, and Isabelle Augenstein. 2022.
\newblock \href {https://doi.org/10.3390/info13100500} {Generating fluent fact
  checking explanations with unsupervised post-editing}.
\newblock \emph{Information}, 13(10):500.

\bibitem[{Kojima et~al.(2022)Kojima, Gu, Reid, Matsuo, and
  Iwasawa}]{DBLP:journals/corr/abs-2205-11916}
Takeshi Kojima, Shixiang~Shane Gu, Machel Reid, Yutaka Matsuo, and Yusuke
  Iwasawa. 2022.
\newblock \href {https://doi.org/10.48550/arXiv.2205.11916} {Large language
  models are zero-shot reasoners}.
\newblock \emph{CoRR}, abs/2205.11916.

\bibitem[{Kotonya and Toni(2020)}]{DBLP:conf/emnlp/KotonyaT20}
Neema Kotonya and Francesca Toni. 2020.
\newblock \href {https://doi.org/10.18653/v1/2020.emnlp-main.623} {Explainable
  automated fact-checking for public health claims}.
\newblock In \emph{Proceedings of the 2020 Conference on Empirical Methods in
  Natural Language Processing (EMNLP)}, pages 7740--7754, Online.

\bibitem[{Krishna et~al.(2022)Krishna, Riedel, and
  Vlachos}]{DBLP:journals/tacl/Krishna0022}
Amrith Krishna, Sebastian Riedel, and Andreas Vlachos. 2022.
\newblock \href {https://transacl.org/ojs/index.php/tacl/article/view/3527}
  {{ProoFVer}: Natural logic theorem proving for fact verification}.
\newblock \emph{Transactions of the Association for Computational Linguistics
  (TACL)}, 10:1013--1030.

\bibitem[{Lee et~al.(2021)Lee, Bang, Madotto, and
  Fung}]{DBLP:conf/naacl/LeeBMF21}
Nayeon Lee, Yejin Bang, Andrea Madotto, and Pascale Fung. 2021.
\newblock \href {https://doi.org/10.18653/v1/2021.naacl-main.158} {Towards
  few-shot fact-checking via perplexity}.
\newblock In \emph{Proceedings of the 2021 Conference of the North American
  Chapter of the Association for Computational Linguistics: Human Language
  Technologies (NAACL-HLT)}, pages 1971--1981, Online.

\bibitem[{Lee et~al.(2020)Lee, Li, Wang, Yih, Ma, and
  Khabsa}]{DBLP:LM_as_fact_checker}
Nayeon Lee, Belinda~Z. Li, Sinong Wang, Wen-tau Yih, Hao Ma, and Madian Khabsa.
  2020.
\newblock \href {https://doi.org/10.18653/v1/2020.fever-1.5} {Language models
  as fact checkers?}
\newblock In \emph{Proceedings of the Third Workshop on Fact Extraction and
  VERification (FEVER)}, pages 36--41, Online.

\bibitem[{Lin et~al.(2021)Lin, Ma, Lin, Yang, Pradeep, and
  Nogueira}]{DBLP:conf/sigir/LinMLYPN21}
Jimmy Lin, Xueguang Ma, Sheng{-}Chieh Lin, Jheng{-}Hong Yang, Ronak Pradeep,
  and Rodrigo Nogueira. 2021.
\newblock \href {https://arxiv.org/abs/2102.10073} {Pyserini: {A} {P}ython
  toolkit for reproducible information retrieval research with sparse and dense
  representations}.
\newblock In \emph{Proceedings of the 44th International {ACM} {SIGIR}
  Conference on Research and Development in Information Retrieval (SIGIR)},
  pages 2356--2362, Online.

\bibitem[{Liu et~al.(2022)Liu, Swayamdipta, Smith, and
  Choi}]{DBLP:journals/corr/abs-2201-05955}
Alisa Liu, Swabha Swayamdipta, Noah~A. Smith, and Yejin Choi. 2022.
\newblock \href {https://aclanthology.org/2022.findings-emnlp.508} {{WANLI}:
  Worker and {AI} collaboration for natural language inference dataset
  creation}.
\newblock In \emph{Findings of the Association for Computational Linguistics:
  EMNLP 2022}, pages 6826--6847, Abu Dhabi, United Arab Emirates.

\bibitem[{Liu et~al.(2019)Liu, Ott, Goyal, Du, Joshi, Chen, Levy, Lewis,
  Zettlemoyer, and Stoyanov}]{DBLP:journals/corr/abs-1907-11692}
Yinhan Liu, Myle Ott, Naman Goyal, Jingfei Du, Mandar Joshi, Danqi Chen, Omer
  Levy, Mike Lewis, Luke Zettlemoyer, and Veselin Stoyanov. 2019.
\newblock \href {https://arxiv.org/abs/1907.11692} {{RoBERTa}: {A} robustly
  optimized {BERT} pretraining approach}.
\newblock \emph{ArXiv preprint}, abs/1907.11692.

\bibitem[{Liu et~al.(2020)Liu, Xiong, Sun, and Liu}]{DBLP:conf/acl/LiuXSL20}
Zhenghao Liu, Chenyan Xiong, Maosong Sun, and Zhiyuan Liu. 2020.
\newblock \href {https://doi.org/10.18653/v1/2020.acl-main.655} {Fine-grained
  fact verification with kernel graph attention network}.
\newblock In \emph{Proceedings of the 58th Annual Meeting of the Association
  for Computational Linguistics (ACL)}, pages 7342--7351, Online.

\bibitem[{Lu and Li(2020)}]{DBLP:conf/acl/LuL20}
Yi-Ju Lu and Cheng-Te Li. 2020.
\newblock \href {https://doi.org/10.18653/v1/2020.acl-main.48} {{GCAN}:
  Graph-aware co-attention networks for explainable fake news detection on
  social media}.
\newblock In \emph{Proceedings of the 58th Annual Meeting of the Association
  for Computational Linguistics (ACL)}, pages 505--514, Online.

\bibitem[{Mialon et~al.(2023)Mialon, Dess{\`{\i}}, Lomeli, Nalmpantis,
  Pasunuru, Raileanu, Rozi{\`{e}}re, Schick, Dwivedi{-}Yu, Celikyilmaz, Grave,
  LeCun, and Scialom}]{DBLP:journals/corr/abs-2302-07842}
Gr{\'{e}}goire Mialon, Roberto Dess{\`{\i}}, Maria Lomeli, Christoforos
  Nalmpantis, Ramakanth Pasunuru, Roberta Raileanu, Baptiste Rozi{\`{e}}re,
  Timo Schick, Jane Dwivedi{-}Yu, Asli Celikyilmaz, Edouard Grave, Yann LeCun,
  and Thomas Scialom. 2023.
\newblock \href {https://doi.org/10.48550/arXiv.2302.07842} {Augmented language
  models: a survey}.
\newblock \emph{CoRR}, abs/2302.07842.

\bibitem[{Nakov et~al.(2022)Nakov, Barr\'{o}n-Cede\~{n}o, Da~San~Martino, Alam,
  Stru\ss{}, Mandl, M\'{\i}guez, Caselli, Kutlu, Zaghouani, Li, Shaar, Shahi,
  Mubarak, Nikolov, Babulkov, Kartal, and Beltr\'{a}n}]{ECIR:CLEF:2022}
Preslav Nakov, Alberto Barr\'{o}n-Cede\~{n}o, Giovanni Da~San~Martino, Firoj
  Alam, Julia~Maria Stru\ss{}, Thomas Mandl, Rub\'{e}n M\'{\i}guez, Tommaso
  Caselli, Mucahid Kutlu, Wajdi Zaghouani, Chengkai Li, Shaden Shaar,
  Gautam~Kishore Shahi, Hamdy Mubarak, Alex Nikolov, Nikolay Babulkov,
  Yavuz~Selim Kartal, and Javier Beltr\'{a}n. 2022.
\newblock \href {https://doi.org/10.1007/978-3-030-99739-7_52} {The {CLEF-2022
  CheckThat!} lab on fighting the {COVID-19} infodemic and fake news
  detection}.
\newblock In \emph{Proceedings of the 44th European Conference on IR Research:
  Advances in Information Retrieval (ECIR)}, pages 416--428, Berlin,
  Heidelberg.

\bibitem[{Nakov et~al.(2021{\natexlab{a}})Nakov, Corney, Hasanain, Alam,
  Elsayed, Barrón-Cedeño, Papotti, Shaar, and
  Da~San~Martino}]{ijcai2021p0619}
Preslav Nakov, David Corney, Maram Hasanain, Firoj Alam, Tamer Elsayed, Alberto
  Barrón-Cedeño, Paolo Papotti, Shaden Shaar, and Giovanni Da~San~Martino.
  2021{\natexlab{a}}.
\newblock \href {https://doi.org/10.24963/ijcai.2021/619} {Automated
  fact-checking for assisting human fact-checkers}.
\newblock In \emph{Proceedings of the Joint Conference on Artificial
  Intelligence (IJCAI)}, pages 4551--4558, Online.

\bibitem[{Nakov et~al.(2021{\natexlab{b}})Nakov, Da~San~Martino, Elsayed,
  Barr{\'{o}}n{-}Cede{\~{n}}o, M\'{i}guez, Shaar, Alam, Haouari, Hasanain,
  Babulkov, Nikolov, Shahi, Struß, and Mandl}]{CheckThat:ECIR2021}
Preslav Nakov, Giovanni Da~San~Martino, Tamer Elsayed, Alberto
  Barr{\'{o}}n{-}Cede{\~{n}}o, Rub\'{e}n M\'{i}guez, Shaden Shaar, Firoj Alam,
  Fatima Haouari, Maram Hasanain, Nikolay Babulkov, Alex Nikolov,
  Gautam~Kishore Shahi, Julia~Maria Struß, and Thomas Mandl.
  2021{\natexlab{b}}.
\newblock \href
  {https://link.springer.com/chapter/10.1007/978-3-030-72240-1_75} {The
  {CLEF}-2021 {CheckThat}! lab on detecting check-worthy claims, previously
  fact-checked claims, and fake news}.
\newblock In \emph{Proceedings of the 43rd European Conference on Information
  Retrieval (ECIR)}, pages 639--649, Lucca, Italy.

\bibitem[{Nguyen et~al.(2020)Nguyen, Sugiyama, Nakov, and Kan}]{FANG}
Van{-}Hoang Nguyen, Kazunari Sugiyama, Preslav Nakov, and Min{-}Yen Kan. 2020.
\newblock \href {https://doi.org/10.1145/3340531.3412046} {{FANG:} leveraging
  social context for fake news detection using graph representation}.
\newblock In \emph{Proceedings of the 29th {ACM} International Conference on
  Information and Knowledge Management (CIKM)}, pages 1165--1174.

\bibitem[{Nie et~al.(2019)Nie, Chen, and Bansal}]{DBLP:conf/aaai/NieCB19}
Yixin Nie, Haonan Chen, and Mohit Bansal. 2019.
\newblock \href {https://doi.org/10.1609/aaai.v33i01.33016859} {Combining fact
  extraction and verification with neural semantic matching networks}.
\newblock In \emph{Proceedings of the 33rd {AAAI} Conference on Artificial
  Intelligence (AAAI)}, pages 6859--6866, Honolulu, Hawaii, USA.

\bibitem[{Nie et~al.(2020)Nie, Williams, Dinan, Bansal, Weston, and
  Kiela}]{DBLP:conf/acl/NieWDBWK20}
Yixin Nie, Adina Williams, Emily Dinan, Mohit Bansal, Jason Weston, and Douwe
  Kiela. 2020.
\newblock \href {https://doi.org/10.18653/v1/2020.acl-main.441} {Adversarial
  {NLI}: A new benchmark for natural language understanding}.
\newblock In \emph{Proceedings of the 58th Annual Meeting of the Association
  for Computational Linguistics (ACL)}, pages 4885--4901, Online.

\bibitem[{Ouyang et~al.(2022)Ouyang, Wu, Jiang, Almeida, Wainwright, Mishkin,
  Zhang, Agarwal, Slama, Ray, Schulman, Hilton, Kelton, Miller, Simens, Askell,
  Welinder, Christiano, Leike, and Lowe}]{DBLP:journals/corr/abs-2203-02155}
Long Ouyang, Jeff Wu, Xu~Jiang, Diogo Almeida, Carroll~L. Wainwright, Pamela
  Mishkin, Chong Zhang, Sandhini Agarwal, Katarina Slama, Alex Ray, John
  Schulman, Jacob Hilton, Fraser Kelton, Luke Miller, Maddie Simens, Amanda
  Askell, Peter Welinder, Paul~F. Christiano, Jan Leike, and Ryan Lowe. 2022.
\newblock \href {https://doi.org/10.48550/arXiv.2203.02155} {Training language
  models to follow instructions with human feedback}.
\newblock \emph{CoRR}, abs/2203.02155.

\bibitem[{Pan et~al.(2021)Pan, Chen, Xiong, Kan, and
  Wang}]{DBLP:conf/acl/PanCXKW20}
Liangming Pan, Wenhu Chen, Wenhan Xiong, Min-Yen Kan, and William~Yang Wang.
  2021.
\newblock \href {https://doi.org/10.18653/v1/2021.acl-short.61} {Zero-shot fact
  verification by claim generation}.
\newblock In \emph{Proceedings of the 59th Annual Meeting of the Association
  for Computational Linguistics and the 11th International Joint Conference on
  Natural Language Processing (ACL-IJCNLP)}, pages 476--483, Online.

\bibitem[{Parrish et~al.(2021)Parrish, Huang, Agha, Lee, Nangia, Warstadt,
  Aggarwal, Allaway, Linzen, and Bowman}]{DBLP:conf/emnlp/ParrishHALNWAAL21}
Alicia Parrish, William Huang, Omar Agha, Soo-Hwan Lee, Nikita Nangia, Alexia
  Warstadt, Karmanya Aggarwal, Emily Allaway, Tal Linzen, and Samuel~R. Bowman.
  2021.
\newblock \href {https://doi.org/10.18653/v1/2021.findings-emnlp.421} {Does
  putting a linguist in the loop improve {NLU} data collection?}
\newblock In \emph{Findings of the Association for Computational Linguistics:
  EMNLP 2021}, pages 4886--4901, Punta Cana, Dominican Republic.

\bibitem[{Popat et~al.(2017)Popat, Mukherjee, Str{\"{o}}tgen, and
  Weikum}]{DBLP:conf/www/PopatMSW17}
Kashyap Popat, Subhabrata Mukherjee, Jannik Str{\"{o}}tgen, and Gerhard Weikum.
  2017.
\newblock \href {https://doi.org/10.1145/3041021.3055133} {Where the truth
  lies: Explaining the credibility of emerging claims on the web and social
  media}.
\newblock In \emph{Proceedngs of the International World Wide Web Conference
  (WWW)}, pages 1003--1012.

\bibitem[{Press et~al.(2022)Press, Zhang, Min, Schmidt, Smith, and
  Lewis}]{DBLP:journals/corr/abs-2210-03350}
Ofir Press, Muru Zhang, Sewon Min, Ludwig Schmidt, Noah~A. Smith, and Mike
  Lewis. 2022.
\newblock \href {https://doi.org/10.48550/arXiv.2210.03350} {Measuring and
  narrowing the compositionality gap in language models}.
\newblock \emph{CoRR}, abs/2210.03350.

\bibitem[{Raffel et~al.(2020)Raffel, Shazeer, Roberts, Lee, Narang, Matena,
  Zhou, Li, and Liu}]{DBLP:journals/jmlr/RaffelSRLNMZLL20}
Colin Raffel, Noam Shazeer, Adam Roberts, Katherine Lee, Sharan Narang, Michael
  Matena, Yanqi Zhou, Wei Li, and Peter~J. Liu. 2020.
\newblock \href {http://jmlr.org/papers/v21/20-074.html} {Exploring the limits
  of transfer learning with a unified text-to-text transformer}.
\newblock \emph{J. Mach. Learn. Res.}, 21:140:1--140:67.

\bibitem[{Robertson and Zaragoza(2009)}]{DBLP:journals/ftir/RobertsonZ09}
Stephen~E. Robertson and Hugo Zaragoza. 2009.
\newblock \href {https://doi.org/10.1561/1500000019} {The probabilistic
  relevance framework: {BM25} and beyond}.
\newblock \emph{Foundations and Trends in Information Retrieval},
  3(4):333--389.

\bibitem[{Saakyan et~al.(2021)Saakyan, Chakrabarty, and
  Muresan}]{DBLP:conf/acl/SaakyanCM20}
Arkadiy Saakyan, Tuhin Chakrabarty, and Smaranda Muresan. 2021.
\newblock \href {https://doi.org/10.18653/v1/2021.acl-long.165} {{COVID}-fact:
  Fact extraction and verification of real-world claims on {COVID}-19
  pandemic}.
\newblock In \emph{Proceedings of the 59th Annual Meeting of the Association
  for Computational Linguistics and the 11th International Joint Conference on
  Natural Language Processing (ACL-IJCNLP)}, pages 2116--2129, Online.

\bibitem[{Sathe et~al.(2020)Sathe, Ather, Le, Perry, and
  Park}]{DBLP:conf/lrec/SatheALPP20}
Aalok Sathe, Salar Ather, Tuan~Manh Le, Nathan Perry, and Joonsuk Park. 2020.
\newblock \href {https://aclanthology.org/2020.lrec-1.849} {Automated
  fact-checking of claims from {W}ikipedia}.
\newblock In \emph{Proceedings of the Twelfth Language Resources and Evaluation
  Conference (LREC)}, pages 6874--6882, Marseille, France.

\bibitem[{Schick et~al.(2023)Schick, Dwivedi{-}Yu, Dess{\`{\i}}, Raileanu,
  Lomeli, Zettlemoyer, Cancedda, and
  Scialom}]{DBLP:journals/corr/abs-2302-04761}
Timo Schick, Jane Dwivedi{-}Yu, Roberto Dess{\`{\i}}, Roberta Raileanu, Maria
  Lomeli, Luke Zettlemoyer, Nicola Cancedda, and Thomas Scialom. 2023.
\newblock \href {https://doi.org/10.48550/arXiv.2302.04761} {Toolformer:
  Language models can teach themselves to use tools}.
\newblock \emph{CoRR}, abs/2302.04761.

\bibitem[{Schuster et~al.(2021)Schuster, Fisch, and
  Barzilay}]{DBLP:conf/naacl/SchusterFB21}
Tal Schuster, Adam Fisch, and Regina Barzilay. 2021.
\newblock \href {https://doi.org/10.18653/v1/2021.naacl-main.52} {Get your
  vitamin {C}! robust fact verification with contrastive evidence}.
\newblock In \emph{Proceedings of the 2021 Conference of the North American
  Chapter of the Association for Computational Linguistics: Human Language
  Technologies (NAACL-HLT)}, pages 624--643, Online.

\bibitem[{Soleimani et~al.(2020)Soleimani, Monz, and
  Worring}]{DBLP:conf/ecir/SoleimaniMW20}
Amir Soleimani, Christof Monz, and Marcel Worring. 2020.
\newblock \href {https://doi.org/10.1007/978-3-030-45442-5\_45} {{BERT} for
  evidence retrieval and claim verification}.
\newblock In \emph{Advances in Information Retrieval (ECIR)}, volume 12036,
  pages 359--366.

\bibitem[{Thorne and Vlachos(2018)}]{DBLP:conf/coling/ThorneV18}
James Thorne and Andreas Vlachos. 2018.
\newblock \href {https://aclanthology.org/C18-1283} {Automated fact checking:
  Task formulations, methods and future directions}.
\newblock In \emph{Proceedings of the 27th International Conference on
  Computational Linguistics (COLING)}, pages 3346--3359, Santa Fe, New Mexico,
  USA.

\bibitem[{Thorne et~al.(2018)Thorne, Vlachos, Christodoulopoulos, and
  Mittal}]{DBLP:conf/naacl/ThorneVCM18}
James Thorne, Andreas Vlachos, Christos Christodoulopoulos, and Arpit Mittal.
  2018.
\newblock \href {https://doi.org/10.18653/v1/N18-1074} {{FEVER}: a large-scale
  dataset for fact extraction and {VER}ification}.
\newblock In \emph{Proceedings of the 2018 Conference of the North {A}merican
  Chapter of the Association for Computational Linguistics: Human Language
  Technologies (NAACL-HLT)}, pages 809--819, New Orleans, Louisiana.

\bibitem[{Vaswani et~al.(2017)Vaswani, Shazeer, Parmar, Uszkoreit, Jones,
  Gomez, Kaiser, and Polosukhin}]{DBLP:conf/nips/VaswaniSPUJGKP17}
Ashish Vaswani, Noam Shazeer, Niki Parmar, Jakob Uszkoreit, Llion Jones,
  Aidan~N. Gomez, Lukasz Kaiser, and Illia Polosukhin. 2017.
\newblock \href
  {https://proceedings.neurips.cc/paper/2017/hash/3f5ee243547dee91fbd053c1c4a845aa-Abstract.html}
  {Attention is all you need}.
\newblock In \emph{Advances in Neural Information Processing Systems 30: Annual
  Conference on Neural Information Processing Systems (NeurIPS)}, pages
  5998--6008, Long Beach, California, USA.

\bibitem[{Wadden et~al.(2020)Wadden, Lin, Lo, Wang, van Zuylen, Cohan, and
  Hajishirzi}]{DBLP:conf/emnlp/WaddenLLWZCH20}
David Wadden, Shanchuan Lin, Kyle Lo, Lucy~Lu Wang, Madeleine van Zuylen, Arman
  Cohan, and Hannaneh Hajishirzi. 2020.
\newblock \href {https://doi.org/10.18653/v1/2020.emnlp-main.609} {Fact or
  fiction: Verifying scientific claims}.
\newblock In \emph{Proceedings of the 2020 Conference on Empirical Methods in
  Natural Language Processing (EMNLP)}, pages 7534--7550, Online.

\bibitem[{Wadden et~al.(2022{\natexlab{a}})Wadden, Lo, Kuehl, Cohan, Beltagy,
  Wang, and Hajishirzi}]{wadden-etal-2022-scifact}
David Wadden, Kyle Lo, Bailey Kuehl, Arman Cohan, Iz~Beltagy, Lucy~Lu Wang, and
  Hannaneh Hajishirzi. 2022{\natexlab{a}}.
\newblock \href {https://aclanthology.org/2022.findings-emnlp.347}
  {{S}ci{F}act-open: Towards open-domain scientific claim verification}.
\newblock In \emph{Findings of the Association for Computational Linguistics:
  EMNLP 2022}, pages 4719--4734, Abu Dhabi, United Arab Emirates.

\bibitem[{Wadden et~al.(2022{\natexlab{b}})Wadden, Lo, Wang, Cohan, Beltagy,
  and Hajishirzi}]{DBLP:conf/naacl/WaddenLWCBH22}
David Wadden, Kyle Lo, Lucy Wang, Arman Cohan, Iz~Beltagy, and Hannaneh
  Hajishirzi. 2022{\natexlab{b}}.
\newblock \href {https://doi.org/10.18653/v1/2022.findings-naacl.6}
  {{M}ulti{V}er{S}: Improving scientific claim verification with weak
  supervision and full-document context}.
\newblock In \emph{Findings of the Association for Computational Linguistics:
  NAACL 2022}, pages 61--76, Seattle, Washington, USA.

\bibitem[{Wang(2017)}]{wang-2017liar}
William~Yang Wang. 2017.
\newblock \href {https://doi.org/10.18653/v1/P17-2067} {{``}{L}iar, liar pants
  on fire{''}: A new benchmark dataset for fake news detection}.
\newblock In \emph{Proceedings of the 55th Annual Meeting of the Association
  for Computational Linguistics (ACL)}, pages 422--426, Vancouver, Canada.

\bibitem[{Wang et~al.(2022)Wang, Wei, Schuurmans, Le, Chi, and
  Zhou}]{DBLP:journals/corr/abs-2203-11171}
Xuezhi Wang, Jason Wei, Dale Schuurmans, Quoc~V. Le, Ed~H. Chi, and Denny Zhou.
  2022.
\newblock \href {https://doi.org/10.48550/arXiv.2203.11171} {Self-consistency
  improves chain of thought reasoning in language models}.
\newblock \emph{CoRR}, abs/2203.11171.

\bibitem[{Wei et~al.(2022)Wei, Wang, Schuurmans, Bosma, Chi, Le, and
  Zhou}]{DBLP:journals/corr/abs-2201-11903}
Jason Wei, Xuezhi Wang, Dale Schuurmans, Maarten Bosma, Ed~H. Chi, Quoc Le, and
  Denny Zhou. 2022.
\newblock \href {https://arxiv.org/abs/2201.11903} {Chain of thought prompting
  elicits reasoning in large language models}.
\newblock \emph{ArXiv preprint}, abs/2201.11903.

\bibitem[{Williams et~al.(2018)Williams, Nangia, and
  Bowman}]{DBLP:conf/naacl/WilliamsNB18}
Adina Williams, Nikita Nangia, and Samuel Bowman. 2018.
\newblock \href {https://doi.org/10.18653/v1/N18-1101} {A broad-coverage
  challenge corpus for sentence understanding through inference}.
\newblock In \emph{Proceedings of the 2018 Conference of the North {A}merican
  Chapter of the Association for Computational Linguistics: Human Language
  Technologies (NAACL-HLT)}, pages 1112--1122, New Orleans, Louisiana, USA.

\bibitem[{Wright et~al.(2022)Wright, Wadden, Lo, Kuehl, Cohan, Augenstein, and
  Wang}]{DBLP:conf/acl/0001WLKCAW22}
Dustin Wright, David Wadden, Kyle Lo, Bailey Kuehl, Arman Cohan, Isabelle
  Augenstein, and Lucy Wang. 2022.
\newblock \href {https://doi.org/10.18653/v1/2022.acl-long.175} {Generating
  scientific claims for zero-shot scientific fact checking}.
\newblock In \emph{Proceedings of the 60th Annual Meeting of the Association
  for Computational Linguistics (ACL)}, pages 2448--2460, Dublin, Ireland.

\bibitem[{Yang et~al.(2019)Yang, Pentyala, Mohseni, Du, Yuan, Linder, Ragan,
  Ji, and Hu}]{DBLP:conf/www/YangPMDYLRJH19}
Fan Yang, Shiva~K. Pentyala, Sina Mohseni, Mengnan Du, Hao Yuan, Rhema Linder,
  Eric~D. Ragan, Shuiwang Ji, and Xia~(Ben) Hu. 2019.
\newblock \href {https://doi.org/10.1145/3308558.3314119} {{XFake}: Explainable
  fake news detector with visualizations}.
\newblock In \emph{Proceedings of the The World Wide Web Conference (WWW)},
  pages 3600--3604, San Francisco, California, USA.

\bibitem[{Yang et~al.(2018)Yang, Qi, Zhang, Bengio, Cohen, Salakhutdinov, and
  Manning}]{DBLP:conf/emnlp/Yang0ZBCSM18}
Zhilin Yang, Peng Qi, Saizheng Zhang, Yoshua Bengio, William Cohen, Ruslan
  Salakhutdinov, and Christopher~D. Manning. 2018.
\newblock \href {https://doi.org/10.18653/v1/D18-1259} {{H}otpot{QA}: A dataset
  for diverse, explainable multi-hop question answering}.
\newblock In \emph{Proceedings of the 2018 Conference on Empirical Methods in
  Natural Language Processing (EMNLP)}, pages 2369--2380, Brussels, Belgium.

\bibitem[{Zhong et~al.(2020)Zhong, Xu, Tang, Xu, Duan, Zhou, Wang, and
  Yin}]{DBLP:conf/acl/ZhongXTXDZWY20}
Wanjun Zhong, Jingjing Xu, Duyu Tang, Zenan Xu, Nan Duan, Ming Zhou, Jiahai
  Wang, and Jian Yin. 2020.
\newblock \href {https://doi.org/10.18653/v1/2020.acl-main.549} {Reasoning over
  semantic-level graph for fact checking}.
\newblock In \emph{Proceedings of the 58th Annual Meeting of the Association
  for Computational Linguistics (ACL)}, pages 6170--6180, Online.

\bibitem[{Zhou et~al.(2019)Zhou, Han, Yang, Liu, Wang, Li, and
  Sun}]{DBLP:conf/acl/ZhouHYLWLS19}
Jie Zhou, Xu~Han, Cheng Yang, Zhiyuan Liu, Lifeng Wang, Changcheng Li, and
  Maosong Sun. 2019.
\newblock \href {https://doi.org/10.18653/v1/P19-1085} {{GEAR}: Graph-based
  evidence aggregating and reasoning for fact verification}.
\newblock In \emph{Proceedings of the 57th Annual Meeting of the Association
  for Computational Linguistics (ACL)}, pages 892--901, Florence, Italy.

\end{thebibliography}
\bibliographystyle{acl_natbib}

\clearpage
\appendix

\section{Implementation Details about the Baselines}
\label{appendix:baselines}

In this section, we give the implementation details for the seven baselines we used in our work. Typical ways to perform few-shot fact-checking using large language models are fine-tuning and in-context learning. Thus, we categorize the baselines into three categories. 

\subsection{Pre-trained Models} 

Pre-trained models use pretrained Transformers~\cite{DBLP:conf/nips/VaswaniSPUJGKP17} such as BERT~\cite{DBLP:conf/naacl/DevlinCLT19} and T5~\cite{DBLP:journals/jmlr/RaffelSRLNMZLL20} for fact-checking. For few-shot learning, we fine-tune them using 20 randomly sampled training examples from HOVER or FEVEROUS. We ran the training 10 times with different random seeds and report the average performance on the validation set. We chose two models: 

\begin{itemize}
    \item \texttt{BERT-FC}~\cite{DBLP:conf/ecir/SoleimaniMW20}: It uses BERT for claim verification. The claim and the evidence are concatenated (\texttt{[CLS] \textit{claim} [SEP] \textit{evidence}}) and used as input for a binary classification task to predict the veracity label of the claim. We use the \texttt{bert-large-uncased} (345M parameters) model provided in HuggingFace.\footnote{\url{https://huggingface.co/}} 
    \item \texttt{LisT5}~\cite{DBLP:conf/acl/JiangPL20}: This is a fact-checking framework built with a pretrained sequence-to-sequence transformer, namely T5~\cite{DBLP:journals/jmlr/RaffelSRLNMZLL20}, as its backbone. We adopt the ``listwise concatenation'' proposed in the paper for label prediction, which concatenates all candidate evidence sentences into a single input and we train the \texttt{t5-large} model to directly classify the claim as \texttt{Supported} or \texttt{Refuted}. We use the original implementation of this model.\footnote{\url{https://github.com/castorini/pygaggle/tree/master/experiments/list5}} 
\end{itemize}

\subsection{FC/NLI Fine-Tuned Models} 
These models are pretrained Transformer models that have been specifically fine-tuned on single-hop fact-checking datasets (\textit{e.g.}, FEVER) or natural language inference (NLI) datasets. This additional training allows these models to excel at fact-checking simple claims, and thus they can generalize better to complex claims that require  multi-hop reasoning during further few-shot fine-tuning. In this category, we selected the following three fine-tuned models:

\begin{itemize}
    \item \texttt{RoBERTa-NLI}~\cite{DBLP:conf/acl/NieWDBWK20} fine-tunes RoBERTa-large~\cite{DBLP:journals/corr/abs-1907-11692} on a combination of four well-known NLI datasets: SNLI~\cite{DBLP:conf/emnlp/BowmanAPM15}, MNLI~\cite{DBLP:conf/naacl/WilliamsNB18}, FEVER-NLI~\cite{DBLP:conf/aaai/NieCB19}, ANLI (R1, R2, R3)~\cite{DBLP:conf/acl/NieWDBWK20}. We used the public model checkpoint available at HuggingFace\footnote{\url{https://huggingface.co/ynie/roberta-large-snli_mnli_fever_anli_R1_R2_R3-nli}} and we further fine-tuned it with 20 random examples from HOVER/FEVEROUS.
    
    \item \texttt{DeBERTaV3-NLI}~\cite{DBLP:journals/corr/abs-2111-09543} fine-tunes the \texttt{DeBERTaV3-large} model on 885,242 NLI hypothesis--premise pairs from FEVER and on four NLI datasets: MNLI, ANLI, LingNLI~\cite{DBLP:conf/emnlp/ParrishHALNWAAL21}, and WANLI~\cite{DBLP:journals/corr/abs-2201-05955}. This is the best-performing NLI model on HuggingFace as of 06/06/2022.\footnote{\url{https://huggingface.co/MoritzLaurer/DeBERTa-v3-large-mnli-fever-anli-ling-wanli}}

    \item \texttt{MULTIVERS}~\cite{DBLP:conf/naacl/WaddenLWCBH22}, formerly known as \texttt{LongChecker}, uses the LongFormer~\cite{DBLP:journals/corr/abs-2004-05150} for claim verification to address the long input evidence problem.  We use a model checkpoint fine-tuned on FEVER.\footnote{\url{https://github.com/dwadden/multivers}}
\end{itemize}

\subsection{In-Context Learning Models} 
These models have recently shown strong few-shot learning ability in various NLP tasks. By prompting a large language model with a few in-context examples, the model can quickly learn a task from demonstrations. To make a fair comparison to our model, we choose two in-context learning baselines as follows. 

\begin{itemize}
    \item \texttt{Codex}~\cite{DBLP:journals/corr/abs-2107-03374} is used in our model to generate reasoning programs. One straightforward baseline directly uses it for fact-checking. To this end, we prompt Codex (\texttt{code-davinci-002}) as follows: ``\texttt{<Evidence> Based on the above information, is it true that <Claim>? True or False? The answer is:}''. We prefix the same 20 in-context examples for our model before the prompt as demonstrations. 
    \item \texttt{FLAN-T5}~\cite{DBLP:journals/corr/abs-2210-11416} is an improved version of T5, which is fine-tuned on 1.8K tasks phrased as instructions, with and without exemplars, \textit{i.e.},~zero-shot and few-shot. The model has shown strong performance in various in-context few-shot learning NLP tasks, such as reasoning, and question-answering. We prompt the model with the same format as we used in Section~\ref{sec:substask_functions}: ``\texttt{<Evidence> Q: <Claim> Is it true that <Claim>? True or False? The answer is:}'', prefixing with the same 20 in-context examples. We also use the same model size (\texttt{FLAN-T5-XXL} 3B) with our model for fair comparison. 
\end{itemize}

\section{Examples of Generated Reasoning Programs}
\label{appendix:correct_examples}

\begin{figure*}[!t]
	\centering
	\includegraphics[width=15cm]{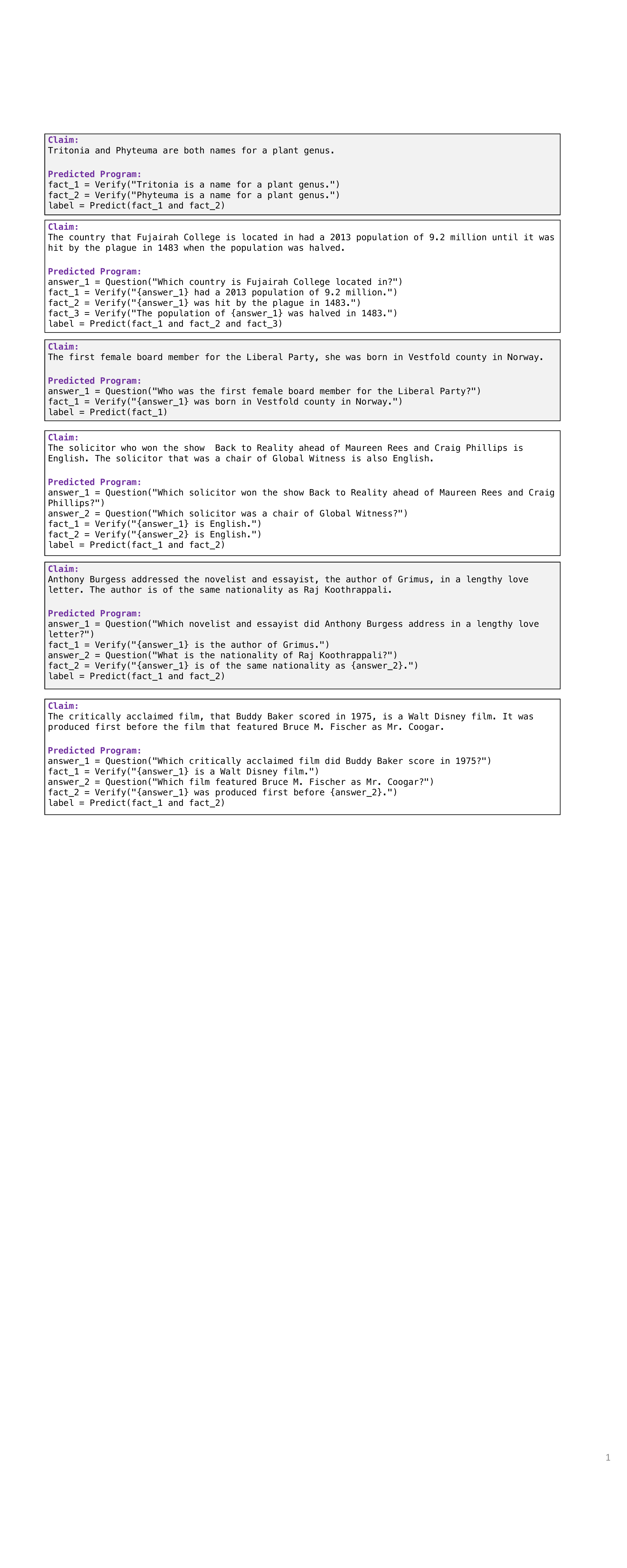}
    \caption{Examples of generated reasoning programs by \model.}
    \label{fig:correct_examples}
\end{figure*}

Figure~\ref{fig:correct_examples} shows six examples of generated reasoning programs by \model that cover diverse reasoning chains. 

\section{Error Analysis for Reasoning Programs}
\label{appendix:examples_programs}

\begin{figure*}[!t]
	\centering
	\includegraphics[width=15cm]{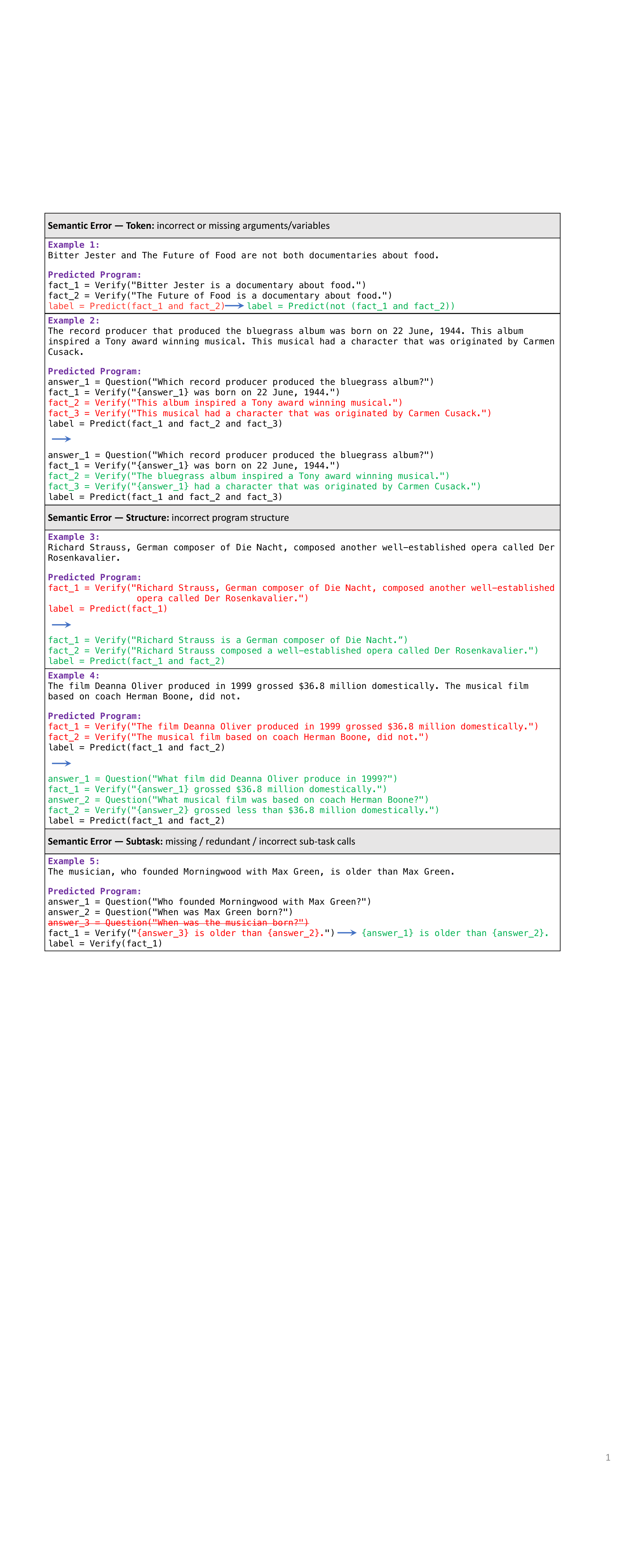}
    \caption{Example error cases where the generated reasoning programs are incorrect. The incorrect segment(s) are marked in {\color{red} \bf red}, and the correct revisions are marked in {\color{darkgreen} \bf green}.}
    \label{fig:error_examples}
\end{figure*}

Figure~\ref{fig:error_examples} shows five examples of erroneous cases where the generated reasoning programs are incorrect. We provide explanations for each of the error cases below:

\paragraph{Example 1} It generates a wrong logical reasoning operator for the final step. The correct logic should be ``\texttt{not (fact\_1 and fact\_2)}'' instead of ``\texttt{fact\_1 and fact\_2}''.

\paragraph{Example 2} It fails to perform co-reference resolution for the arguments in the third and the fourth reasoning steps. ``This album'' should be replaced with ``The bluegrass'' to make the sub-task context-independent. ``This musical'' should be replaced with the variable ``\texttt{answer\_1}'' from the first step.

\paragraph{Example 3} It fails to create a meaningful problem decomposition for the claim. It generates a trivial program that simply repeats the original claim.

\paragraph{Example 4} It fails to generate a fine-grained reasoning structure for the input claim. It also generates a trivial program that simply separates the claim into sentences.

\paragraph{Example 5} It generates a redundant reasoning step ``\texttt{Question("When was the musician born?")}'', which does not add any new information to the reasoning chain.

\section{Program Generation Prompts}
\label{appendix:programs}

Our manually written prompts for the HOVER and the FEVEROUS-S datasets are given in Listings~\ref{program_HOVER} and~\ref{program_FEVEROUS}, respectively. 

\section{Prompts for Closed-Book Fact-Checking}
\label{appendix:closed_book_prompt}

Below we show the templates for the four prompting methods used for InstructGPT for the closed-book fact-checking setting in Section~\ref{sec:close_book}. 

\vspace{0.1cm}

\textbf{Direct Prompting}

\lstset{
    style=mystyle,
    basicstyle=\ttfamily\scriptsize,
    backgroundcolor=\color{white},
    stringstyle=\color{black},
    keywordstyle=\color{black},
    breaklines=false,
    keepspaces=false
}
\begin{lstlisting}[language=Python]
# Answer the following true/false questions:

Is it true that The woman the story behind Girl Crazy 
is credited to is older than Ted Kotcheff?
The answer is: False

(*@\color{codegray}{\textbf{($\cdots$ more in-context examples here $\cdots$)}}@*)

Is it true that (*@{\color{codepurple}{\textsc{<input\_claim>}}@*)?
The answer is: 
\end{lstlisting}

\textbf{ZS-CoT Prompting}

\begin{lstlisting}[language=Python]
# Answer the following true/false question:

Is it true that (*@{\color{codepurple}{\textsc{<input\_claim>}}@*)? True or False?
Let us think step-by-step. The answer is: 
\end{lstlisting}

\textbf{CoT Prompting}

\begin{lstlisting}[language=Python]
# Answer the following true/false questions:

Is it true that The woman the story behind Girl Crazy 
is credited to is older than Ted Kotcheff?
Let's think step by step. 
Girl Crazy's story is credited to Hampton Del Ruth. 
Hampton Del Ruth was born on September 7, 1879. 
Ted Kotcheff was born on April 7, 1931. 
Therefore, the answer is: False.

(*@\color{codegray}{\textbf{($\cdots$ more in-context examples here $\cdots$)}}@*)

Is it true that (*@{\color{codepurple}{\textsc{<input\_claim>}}@*)?
Let's think step by step. 
\end{lstlisting}

\textbf{Self-Ask Prompting}

\begin{lstlisting}[language=Python]
# Answer the following true/false questions:

Is it true that The woman the story behind Girl Crazy 
is credited to is older than Ted Kotcheff?
Q: The story behind Girl Crazy is credited to whom?
A: Hampton Del Ruth
Q: Is Hampton Del Ruth older than Ted Kotcheff?
A: No
So the final answer is: False.

(*@\color{codegray}{\textbf{($\cdots$ more in-context examples here $\cdots$)}}@*)

Is it true that (*@{\color{codepurple}{\textsc{<input\_claim>}}@*)?
\end{lstlisting}

\clearpage

\onecolumn
\lstset{
    style=mystyle,
    caption=The prompt used for Program Generation for HOVER.,
    label=program_HOVER
}
\begin{lstlisting}[language=Python]
'''Generate a python-like program that describes the reasoning steps required to verify the claim step-by-step. You can call three functions in the program: 1. Question () to answer a question; 2. Verify () to verify a simple claim; 3. Predict() to predict the veracity label.'''

# The claim is that Howard University Hospital and Providence Hospital are both located in Washington, D.C.
def program():
    fact_1 = Verify("Howard University Hospital is located in Washington, D.C.")
    fact_2 = Verify("Providence Hospital is located in Washington, D.C.")
    label = Predict(fact_1 and fact_2)

# The claim is that WWE Super Tuesday took place at an arena that currently goes by the name TD Garden.
def program():
    answer_1 = Question("Which arena the WWE Super Tuesday took place?")
    fact_1 = Verify(f"{answer_1} currently goes by the name TD Garden.")
    label = Predict(fact_1)

# The claim is that Talking Heads, an American rock band that was "one of the most critically acclaimed bands of the 80's" is featured in KSPN's AAA format.
def program():
    fact_1 = Verify("Talking Heads is an American rock band that was 'one of the most critically acclaimed bands of the 80's'.")
    fact_2 = Verify("Talking Heads is featured in KSPN's AAA format.")
    label = Predict(fact_1 and fact_2)

# The claim is that An IndyCar race driver drove a Formula 1 car designed by Peter McCool during the 2007 Formula One season.
def program():
    answer_1 = Question("Which Formula 1 car was designed by Peter McCool during the 2007 Formula One season?")
    fact_1 = Verify(f"An IndyCar race driver drove the car {answer_1}.")
    label = Predict(fact_1)
    
# The claim is that Gina Bramhill was born in a village. The 2011 population of the area that includes this village was 167,446.
def program():
    answer_1 = Question("Which village was Gina Bramhill born in?")
    fact_1 = Verify(f"The 2011 population of the area that includes {answer_1} was 167,446.")
    label = Predict(fact_1)
    
# The claim is that Don Ashley Turlington graduated from Saint Joseph's College, a private Catholic liberal arts college in Standish.
def program():
    fact_1 = Verify("Saint Joseph's College is a private Catholic liberal arts college is located in Standish.")
    fact_2 = Verify(f"Don Ashley Turlington graduated from Saint Joseph's College.")
    label = Predict(fact_1 and fact_2)
    
# The claim is that Gael and Fitness are not published in the same country.
def program():
    answer_1 = Question("Which country was Gael published in?")
    answer_2 = Question("Which country was Fitness published in?")
    fact_1 = Verify(f"{answer_1} and {answer_2} are not the same country.")
    label = Predict(fact_1)

# The claim is that Blackstar is the name of the album released by David Bowie that was recorded in secret.
def program():
    fact_1 = Verify("David Bowie released an album called Blackstar.")
    fact_2 = Verify("David Bowie recorded an album in secret.")
    label = Predict(fact_1 and fact_2)
    
# The claim is that In the 2004 Hockey film produced by a former major league baseball pitcher Kurt Russell played the USA coach.
def program():
    answer_1 = Question("Which 2004 Hockey film was produced a former major league baseball pitcher?")
    fact_1 = Verify("Kurt Russell played the USA coach in the film {answer_1}.")
    label = Predict(fact_1)
    
# The claim is that Along with the New York Islanders and the New York Rangers, the New Jersey Devils NFL franchise is popular in the New York metropolitan area.
def program():
    fact_1 = Verify("The New York Islanders and the New York Rangers are popular in the New York metropolitan area.")
    fact_2 = Verify("The New Jersey Devils NFL franchise is popular in the New York metropolitan area.")
    label = Predict(fact_1 and fact_2)
    
# The claim is that Jack McFarland is the best known role of the host of the 64th Annual Tony Awards.
def program():
    answer_1 = Question("Who is the host of the 64th Annual Tony Awards?")
    fact_1 = Verify(f\"Jack McFarland is the best known role of {answer_1}.")
    label = Predict(fact_1)
    
# The claim is that The song recorded by Fergie that was produced by Polow da Don and was followed by Life Goes On was M.I.L.F.$.
def program():
    fact_1 = Verify("M.I.L.F.$ was recorded by Fergie that was produced by Polow da Don.")
    fact_2 = Verify("M.I.L.F.$ was was followed by Life Goes On.")
    label = Predict(fact_1 and fact_2)

# The claim is that Eatza Pizza and Your Pie were not founded in the same state.
def program():
    answer_1 = Question("Which state was Eatza Pizza founded in?")
    answer_2 = Question("Which state was Your Pie founded in?")
    fact_1 = Verify(f"{answer_1} and {answer_2} are not the same state.")
    label = Predict(fact_1)

# The claim is that Gregg Rolie and Rob Tyner, are not a keyboardist.
def program():
    fact_1 = Verify("Gregg Rolie is not a keyboardist.")
    fact_2 = Verify("Rob Tyner is not a keyboardist.")
    label = Predict(fact_1 and fact_2)
    
# The claim is that Maria Esther Andion Bueno, not Jimmy Connors, is the player that is from Brazil.
def program():
    fact_1 = Verify("Maria Esther Andion Bueno is from Brazil.")
    fact_2 = Verify("Jimmy Connors is not from Brazil.")
    label = Predict(fact_1 and fact_2)
    
# The claim is that Vladimir Igorevich Arnold died after Georg Cantor.
def program():
    answer_1 = Question("When did Vladimir Igorevich Arnold die?")
    answer_2 = Question("When did Georg Cantor die?")
    fact_1 = Verify(f"{answer_1} is after {answer_2}.")
    label = Predict(fact_1)

# The claim is that Barton Mine was halted by a natural disaster not Camlaren Mine.
def program():
    fact_1 = Verify("Barton Mine was halted by a natural disaster.")
    fact_2 = Verify("Camlaren Mine was not halted by a natural disaster.")
    label = Predict(fact_1 and fact_2)
    
# The claim is that John O'Hara and Rabindranath Tagore are not the same nationality.
def program():
    answer_1 = Question("What is the nationality of John O'Hara?")
    answer_2 = Question("What is the nationality of Rabindranath Tagore?")
    fact_1 = Verify(f"{answer_1} and {answer_2} are not the same nationality.")
    label = Predict(fact_1)


# The claim is that Thomas Loren Friedman has won more Pulitzer Prizes than Colson Whitehead.
def program():
    answer_1 = Question("How many Pulitzer Prizes has Thomas Loren Friedman won?")
    answer_2 = Question("How many Pulitzer Prizes has Colson Whitehead won?")
    fact_1 = Verify(f"{answer_1} is more than {answer_2}.")
    label = Predict(fact_1)

# The claim is that The model of car Trevor Bayne drives was introduced for model year 2006. The Rookie of The Year in the 1997 CART season drives it in the NASCAR Sprint Cup.
def program():
    answer_1 = Question("Which model of car is drived by Trevor Bayne?")
    fact_1 = Verify(f"{answer_1} was introduced for model year 2006.")
    answer_2 = Question("Who is the Rookie of The Year in the 1997 CART season?")
    fact_2 = Verify(f"{answer_2} drives the model of car Trevor Bayne drives in the NASCAR Sprint Cup.")
    label = predict(fact_1 and fact_2)
    
# The claim is that (*@{\color{codepurple}{\textsc{<input\_claim>}}@*)
def program():

\end{lstlisting}

\onecolumn
\lstset{
    style=mystyle,
    caption=The prompt used for Program Generation for FEVEROUS-S.,
    label=program_FEVEROUS
}
\begin{lstlisting}[language=Python]

'''Generate a python-like program that describes the reasoning steps required to verify the claim step-by-step. You can call three functions in the program: 1. Question () to answer a question; 2. Verify () to verify a simple claim; 3. Predict() to predict the veracity label.'''

# The claim is that In 1959, former Chilean boxer Alfredo Cornejo Cuevas (born June 6, 1933) won the gold medal in the welterweight division at the Pan American Games (held in Chicago, United States, from August 27 to September 7) in Chicago, United States, and the world amateur welterweight title in Mexico City.
def program():
    fact_1 = Verify("Alfredo Cornejo Cuevas was born in June 6, 1933.")
    fact_2 = Verify("Alfredo Cornejo Cuevas won the gold medal in the welterweight division at the Pan American Games in 1959.")
    fact_3 = Verify("The Pan American Games in 1959 was held in Chicago, United States, from August 27 to September 7.")
    fact_4 = Verify("Alfredo Cornejo Cuevas won the world amateur welterweight title in Mexico City.")
    label = Predict(fact_1 and fact_2 and fact_3 and fact_4)

# The claim is that The Footwork FA12, which was intended to start the season, finally debuted at the San Marino Grand Prix, a Formula One motor race held at Imola on 28 April 1991.
def program():
    fact_1 = Verify("The Footwork FA12, which was intended to start the season.")
    fact_2 = Verify("The Footwork FA12 finally debuted at the San Marino Grand Prix.")
    fact_3 = Verify("The San Marino Grand Prix was a Formula One motor race held at Imola on 28 April 1991.")
    label = Predict(fact_1 and fact_2 and fact_3)

# The claim is that SkyHigh Mount Dandenong (formerly Mount Dandenong Observatory) is a restaurant located on top of Mount Dandenong, Victoria, Australia.
def program():
    fact_1 = Verify("SkyHigh Mount Dandenong is a restaurant located on top of Mount Dandenong, Victoria, Australia.")
    fact_2 = Verify("SkyHigh Mount Dandenong is formerly known as Mount Dandenong Observatory.")
    label = Predict(fact_1 and fact_2)

# The claim is that Before the first Europeans arrived or copra companies leased it, Maupihaa was home to Inca's in ancient times.
def program():
    fact_1 = Verify("Maupihaa was home to Inca's in ancient times.")
    fact_2 = Verify("Maupihaa was home to Inca's before the first Europeans arrived or copra companies leased it.")
    label = Predict(fact_1 and fact_2)
    
# The claim is that Shulin, a 33.1288 km (12.7911 sq mi) land located in New Taipei City, China, a country in East Asia, has a total population of 183,946 in December 2018.
def program():
    fact_1 = Verify("Shulin is a 33.1288 km (12.7911 sq mi) land located in New Taipei City, China.")
    fact_2 = Verify("Shulin has a total population of 183,946 in December 2018.")
    label = Predict(fact_1 and fact_2)
    
# The claim is that Sumo wrestler Toyozakura Toshiaki committed match-fixing, ending his career in 2011 that started in 1989.
def program():
    fact_1 = Verify("Toyozakura Toshiaki ended his career in 2011 that started in 1989.")
    fact_2 = Verify("Toyozakura Toshiaki is a Sumo wrestler.")
    fact_3 = Verify("Toyozakura Toshiaki committed match-fixing.")
    label = Predict(fact_1 and fact_2 and fact_3)

# The claim is that In 1959, former Chilean boxer Alfredo Cornejo Cuevas (born June 6, 1933) won the gold medal in the welterweight division at the Pan American Games (held in Chicago, United States, from August 27 to September 7) in Chicago, United States, and the world amateur welterweight title in Mexico City.
def program():
    fact_1 = Verify("Alfredo Cornejo Cuevas is a former Chilean boxer.")
    fact_2 = Verify("Alfredo Cornejo won the gold medal in the welterweight division at the Pan American Games.")
    fact_3 = Verify("The Pan American Games was held in Chicago, United States, from August 27 to September 7.")
    fact_4 = Verify("Alfredo Cornejo won the world amateur welterweight title in Mexico City.")
    label = Predict(fact_1 and fact_2 and fact_3 and fact_4)

# The claim is that Adductor hiatus is associated with nine structures, seven of which enter and leave through hiatus.
def program():
    fact_1 = Verify("Adductor hiatus is associated with nine structures.")
    fact_2 = Verify("Seven of the nine structures associated with Adductor hiatus enter and leave through hiatus.")
    label = Predict(fact_1 and fact_2)
    
# The claim is that Ifor Bowen Lloyd was educated at Winchester (an independent boarding school for boys in the British public school tradition) and Exeter College, Oxford where he was a member of the Library Committee of the Oxford Union Society, as well as, received a BA in Modern History in 1924.
def program():
    fact_1 = Verify("Ifor Bowen Lloyd was educated at Winchester and Exeter College, Oxford.")
    fact_2 = Verify("Winchester is an independent boarding school for boys in the British public school tradition.")
    fact_3 = Verify("While at Oxford, Ifor Bowen Lloyd was a member of the Library Committee of the Oxford Union Society.")
    fact_4 = Verify("Ifor Bowen Lloyd received a BA in Modern History in 1924 at Oxford.")
    label = Predict(fact_1 and fact_2 and fact_3 and fact_4)
  
# The claim is that In the 2001 Stanley Cup playoffs Eastern Conference Semifinals Devils' Elias scored and Maple Leafs' left Devils player Scott Neidermayer hurt.
def program():
    fact_1 = Verify("In the 2001 Stanley Cup playoffs Eastern Conference Semifinals Devils' Elias scored.")
    fact_2 = Verify("Maple Leafs' left Devils player Scott Neidermayer hurt.")
    label = Predict(fact_1 and fact_2)
    
# The claim is that Teldenia helena is a moth first described in 1967 by Wilkinson.
def program():
    fact_1 = Verify("Teldenia helena is a moth.")
    fact_2 = Verify("Teldenia helena was first described by Wilkinson in 1967.")
    label = Predict(fact_1 and fact_2)
    
# The claim is that Born December 30, 1974, William Frick was a dark horse candidate in the Maryland House of Delegates appointment process.
def program():
    fact_1 = Verify("William Frick was born in December 30, 1974.")
    fact_2 = Verify("William Frick was a dark horse candidate in the Maryland House of Delegates appointment process.")
    label = Predict(fact_1 and fact_2)

# The claim is that (*@{\color{codepurple}{\textsc{<input\_claim>}}@*)
def program():
\end{lstlisting}


\twocolumn

\end{document}